\documentclass{article}
\usepackage{graphicx}
\usepackage{subcaption}
\usepackage{titlesec}

\usepackage[preprint]{neurips_2026}
\usepackage[utf8]{inputenc} 
\usepackage[T1]{fontenc}    
\usepackage{hyperref}       
\usepackage{url}            
\usepackage{booktabs}       
\usepackage{amsfonts}       
\usepackage{nicefrac}       
\usepackage{tcolorbox}
\usepackage{microtype}      
\usepackage{xcolor}         
\usepackage[table]{xcolor}
\usepackage{wrapfig}
\usepackage{booktabs}
\usepackage{amsmath} 
\usepackage{graphicx}  
\usepackage{multirow}
\newcommand{\cmark}{\checkmark}
\newcommand{\xmark}{$\times$}

\title{EgoExoMem: Cross-View Memory Reasoning over Synchronized Egocentric and Exocentric Videos}

\author{
Ruiping Liu$^{1}$, 
Junwei Zheng$^{1,2}$,
Yufan Chen$^{1,3}$,
Di Wen$^{1}$,
Shaofang Quan$^{1}$,
Chengzhi Wu$^{1}$,\\
\textbf{Jiaming Zhang}$^{4}$,
\textbf{Kailun Yang}$^{4}$,
\textbf{Kunyu Peng}$^{1,5}$\thanks{\textbf{Corresponding author}: kunyu.peng@kit.edu;   \textbf{First author}: ruiping.liu@kit.edu},
\textbf{Rainer Stiefelhagen}$^{1}$\\
$^1$Karlsruhe Institute of Technology
$^2$ETH Zurich
$^3$University of Oxford\\
$^4$Hunan University
$^{5}$INSAIT, Sofia University ``St. Kliment Ohridski''
\\
}

\begin{document}

\maketitle
\begin{center}
    \centering
    \captionsetup{type=figure}
    \begin{subfigure}[t]{0.85\textwidth}
        \centering
        \includegraphics[width=\textwidth]{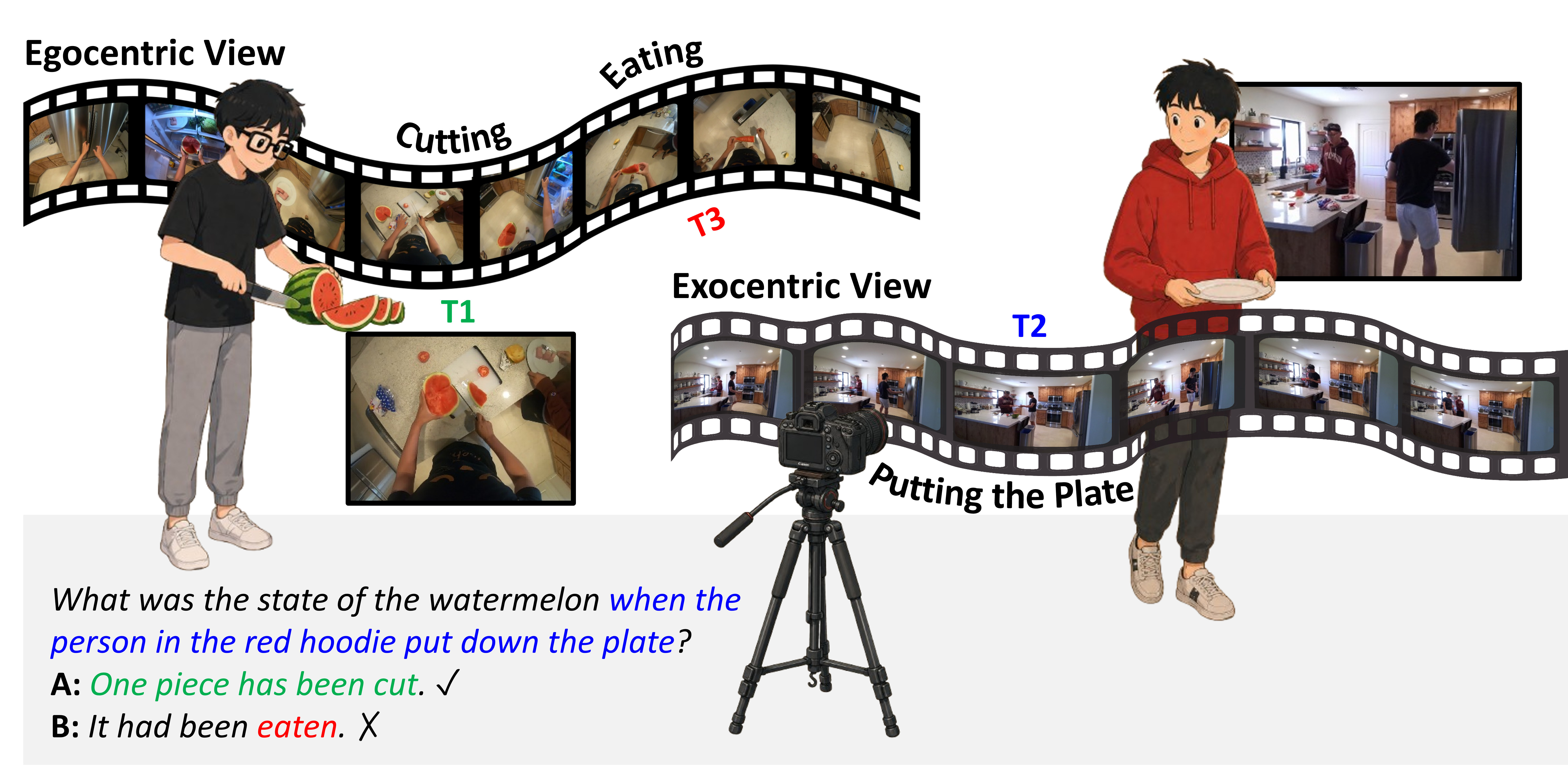}

        \label{fig1-b}
    \end{subfigure}
    \vskip-1.5ex
    \caption{EgoExoMem requires reasoning over synchronized ego-exo memory streams to answer questions unanswerable from either viewpoint alone. \textcolor{blue}{Blue} indicates the temporal condition, \textcolor{green}{green} indicates keyframes or the correct answer, and \textcolor{red}{red} indicates distractors.}
    \label{fig1:banner}
\end{center}%

\begin{abstract}
Egocentric memory is widely used in embodied intelligence, but it may be insufficient for comprehensive spatial-temporal reasoning. Inspired by human recall from both field and observer perspectives, we introduce EgoExoMem, the first benchmark for cross-view memory reasoning over synchronized egocentric and exocentric videos. EgoExoMem contains $2.6K$ high-quality MCQs across eight temporal, spatial, and cross-view QA types. To support dual-view retrieval, we propose E$^2$-Select, a training-free frame selection method for synchronized ego-exo videos. It combines relevance-based budget allocation with per-view k-DPP sampling to handle view asymmetry and cross-view temporal consistency. Experiments show that ego and exo views provide complementary memory cues, while existing MLLMs remain far from solving the benchmark: the best model reaches only $55.3\%$. E$^2$-Select achieves state-of-the-art performance of $58.2\%$ over frame-selection and RAG-based memory baselines. Further analysis reveals systematic view-preference conflicts between question framing and answer grounding, underscoring the novelty and challenge of cross-view memory reasoning. The source code and dataset can be found at \url{https://github.com/RuipingL/EgoExoMem}.
\end{abstract}

\section{Introduction}
With the rapid advancement of multimodal large language models, embodied agents are increasingly expected to perceive, remember, and reason about complex real-world environments~\cite{liu2025aligning}. 
Central to this capability is memory: the ability to retain and retrieve past observations to support informed decision-making and long-horizon task execution~\cite{yadav2024findingdory, fan2025embodied}. 
However, existing memory frameworks for embodied perception are predominantly built upon egocentric observations~\cite{grauman2022ego4d, fan2025embodied}, capturing the world exclusively from the agent's own first-person viewpoint. 
While egocentric memory has been widely studied for tasks such as episodic event recall~\cite{yang2025egolife, ye2025mmego, yeo2025worldmm}, scene understanding~\cite{yang2025thinking, majumdar2023openeqa}, and navigation~\cite{zhou2025learning}, it remains fundamentally limited by the agent's physical vantage point, leaving out-of-view regions and their spatial relationships inconsistently observed. 
This raises a critical question: \textit{is egocentric memory alone sufficient for comprehensive spatial understanding and temporal reasoning in embodied environments?}

Real-world scenarios reveal fundamental limitations of purely egocentric memory. For holistic scene understanding, an agent's egocentric stream captures its own interactions but offers limited visibility into other agents' interactions within the shared space. While equipping all agents with smart glasses has been proposed~\cite{yang2025egolife, kim2026ma}, this is impractical due to device cost and wearable burden~\cite{liu2024objectfinder}. Egocentric recording is also ill-suited for capturing whole-body movements, particularly in healthcare contexts such as fall detection~\cite{yang2024egoposeformer}. Given the growing prevalence of exocentric infrastructure, including hospital surveillance~\cite{schneider2025omnifall, gabriel2025continuous} and smart home cameras~\cite{chen2026mural, ring2024}, we propose complementing egocentric memory with exocentric observations to address these limitations.

Humans naturally encode experiences from two complementary perspectives: field memory, which relives events through a first-person viewpoint, and observer memory, which recalls the same event from a third-person vantage point~\cite{nigro1983pov}.
These perspectives mirror the brain's parallel egocentric and allocentric reference frames~\cite{burgess2006spatial}, and their interplay supports situation awareness~\cite{endsley1995toward}.
Yet, existing ego-exo works focus on imitation learning or unidirectional knowledge transfer~\cite{zhang2026exo2ego}, with methods treating the two streams independently~\cite{huang2024egoexolearn} and benchmarks evaluating only local view-clip matching~\cite{he2025egoexobench} or view-invariant temporal consistency~\cite{jung2025egoexo,reilly2025my}, rather than reasoning over the full, synergistic memory formed by both views.

To address this gap, we introduce \textbf{EgoExoMem, a large-scale benchmark} comprising $2.6K$ human-verified Multiple Choice Questions (MCQs) designed to evaluate memory-based reasoning over synchronized egocentric and exocentric videos sourced from EgoExo4D~\cite{grauman2024ego} and LEMMA~\cite{jia2020lemma}.
We characterize memory-based reasoning along eight QA types covering temporal, spatial, and cross-view dimensions: Habitual Location, Instantaneous Position, Resulting Location, Egocentric Direction, Object State, Allocentric Relation, Third-Person Activity, and Temporal Ordering, with an example illustrated in Fig.~\ref{fig1:banner}. 
A text-only check is applied to ensure the vision-dependency of all answers. Motivated by the view-asymmetric and temporally synchronized nature of EgoExoMem, we further propose \textbf{E$^2$-Select, a simple yet effective frame selection method} with relevance-based budget allocation and k-DPP sampling, explicitly designed to handle view asymmetry and cross-view temporal consistency beyond single-view memory methods.

\textbf{Extensive experiments} are conducted to validate the reasonableness of the EgoExoMem task, benchmark the dataset, demonstrate the effectiveness of E$^2$-Select, and analyze failure cases. Proprietary~\cite{comanici2025gemini} and open-source MLLMs~\cite{li2024llava, wang2025internvl3, bai2025qwen3} are evaluated on EgoExoMem as zero-shot baselines, confirming the task's challenge. Memory mechanisms spanning frame selection~\cite{tang2025adaptive, liu2025bolt}, text retrieval~\cite{robertson2009bm25, karpukhin2020dense}, visually-grounded retrieval~\cite{luo2024video, jeong2025videorag}, and structured memory retrieval~\cite{kim2026ma, yeo2025worldmm} are used to benchmark EgoExoMem, while our E$^2$-Select achieves state-of-the-art performance with an averaged MCQ accuracy of $58.2\%$. Comprehensive ablation studies support the design choices of E$^2$-Select. Failure case analysis reveals a view-preference conflict between question framing and answer grounding, highlighting the complementary potential of synergistic ego-exo views.

\section{Related Work}
\begin{table}[t]
\centering
\caption{Comparison of related video QA benchmarks. ``--'' means not available. }
\label{tab:benchmark_comparison}
\resizebox{\linewidth}{!}{%
\begin{tabular}{lcccccccc}
\toprule
\textbf{Benchmark} & \textbf{Venue} & \textbf{\#QA} & \textbf{\#Videos} & \textbf{\#Tasks} & \textbf{View} & \textbf{Sync} & \textbf{Memory QA} & \textbf{Cross-view QA} \\
\midrule
EgoSchema~\cite{mangalam2023egoschema}       & NeurIPS'23 & 5.1k  & 5.1k & 1  & Ego      & --  & \xmark & \xmark \\
VSI-Bench~\cite{yang2025thinking}            & CVPR'25    & 5.0k  & 288  & 8  & Ego      & --  & \cmark & \xmark \\
EgoMemoria~\cite{ye2025mmego}                  & ICLR'25    & 7.0k  & 629  & -- & Ego     & --  & \cmark & \xmark \\
EgoPlan-Bench~\cite{chen2026egoplan}         & ECCV'24    & 4.9k  & --  & 1  & Ego      & --  & \xmark & \xmark \\
EgoExoLearn~\cite{huang2024egoexolearn}      & CVPR'24    & 2.2k  & 747  & 3  & Ego+Exo  & Async & \xmark & \cmark \\
EgoExoBench~\cite{he2025egoexobench}         & NeurIPS'25 & 7.3k  & --  & 11 & Ego+Exo  & Both  & \xmark & \cmark \\
EPFL-Smart-Kitchen~\cite{bonnetto2025epfl_smart_kitchen} & NeurIPS'25 & 17.7k & 49 & 6 & Ego+Exo & Sync & \cmark & \xmark \\
MA-EgoQA~\cite{kim2026ma}              & arXiv'26   & 1.7k  & --  & 5  & Multi-ego & --  & \xmark & \cmark \\
\midrule
\textbf{EgoExoMem (Ours)}                    & 2026        & 2.6k & 390 & 8 & Ego+Exo & Sync & \cmark & \cmark \\
\bottomrule
\end{tabular}%
}
\label{tab:comparison}
\vskip-2ex
\end{table}
\noindent\textbf{Ego-Exo Video Understanding and Reasoning.}
Cross-view understanding spans view translation~\cite{luo2024put_lifting,zhang2026exo2ego,mahdi2025exo2egosyn,ohkawa2025exo2egodvc,shi2025unsupervised_egoexo}, correspondence learning~\cite{hu2025robust_ego_exo,huang2024egoexolearn}, cross-view grounding~\cite{fu2025objectrelator,su2026regionaligner,luo2024grounded_affordance}, and audio-visual association~\cite{huang2025sound_bridge,jia2024audio_visual}. View-conditioned generation has also advanced rapidly, including ego-exo synthesis~\cite{xu2025egoexo_gen,park2025egoworld,pan2026v2sam}, viewpoint-aware anticipation~\cite{shi2026test_action_anticipation}, and analysis of active versus passive perspectives~\cite{lu2025ego_exo_viewpoint}. 
Beyond generation, view-invariant representation learning~\cite{park2025bootstrap} and imitation error detection~\cite{li2026sava} further demonstrate the value of cross-view alignment for fine-grained activity understanding. 
Recent benchmarks have also expanded the ego-exo scope: EPFL-Smart-Kitchen~\cite{bonnetto2025epfl_smart_kitchen} evaluates action and motion understanding in multi-modal ego-exo settings, and SAW-Bench~\cite{li2026learning_situated_awareness} probes situated awareness in real-world environments. 
Closest to our setting, Ravi~\textit{et al.}~\cite{ravi2025out_sight_context} show that off-screen context can be recovered from egocentric observations, and EgoExoLearn~\cite{huang2024egoexolearn} benchmarks skill assessment across paired viewpoints. 
Yet, existing methods treat the two streams as inputs to a shared representation rather than as asymmetric sources of complementary evidence, a distinction that prior cross-view benchmarks do not evaluate. 
Relevant benchmarks are summarized in Tab.~\ref{tab:comparison}.

\noindent\textbf{Memory-Augmented Video Understanding.}
Episodic memory QA over egocentric video is well established~\cite{barmann2022did_episodic,datta2022episodic_memory,jia2022egotaskqa,jiang2023single_stage,bai2023glance_focus}.
Recent work pushes toward longer temporal horizons~\cite{yang2025egolife,kulkarni2025egovita,fan2025prvql} and richer memory structures, including spatial mind palaces~\cite{ginting2025enter_mind_palace}, dynamic memory refinement~\cite{hu2025cascaded_dynamic_memory}, and multi-agent memory~\cite{mur2025omama, kim2026ma}.
Fine-grained spatiotemporal grounding~\cite{liang2025fine_grained_spatiotemporal,feng2025object_shot,xu2024retrieval_augmented_egocentric} and egocentric captioning~\cite{deichler2025look_tell,kang2023video_captioning} further underpin retrieval capabilities.
However, egocentric streams capture fine-grained interactions but miss the spatial layout and third-person activities. Exocentric streams offer the reverse perspective but lack close-up manipulation detail. EgoExoMem exploits this asymmetry with QA types that are unresolvable from either view alone.

\noindent\textbf{Spatial and Procedural Video QA Benchmarks.}
Spatial QA benchmarks probe object size, distance, room layout~\cite{yang2025thinking}, embodied cognition~\cite{dang2025ecbench}, and broad multi-modal egocentric perception~\cite{luo2025openmmego,peng2025eye,jung2025right}. Procedural and task-oriented QA is addressed across activity sequences~\cite{jia2022egotaskqa}, instructional video settings~\cite{ragusa2026ego_extra}, and long-horizon retrieval tasks~\cite{yang2025egolife,fan2025prvql}. 
Chain-of-thought and cross-modal reasoning in egocentric settings are targeted by Egothinker~\cite{pei2025egothinker} and EgoCross~\cite{li2026egocross}. 
Object-centric and referential understanding~\cite{wang2025object_centric_vqa,li2026mllms_referential,sun2025visual_intention} connects directly to EgoExoMem's localization and relation-change question types, while visual query localization~\cite{jiang2023single_stage} formalizes the \emph{where did I last see X} task that underlies our displacement queries. 
Scene graph understanding~\cite{rodin2025easg,sima2024drivelm,cherian2022_spatio_temporal_scene_graph}, egocentric text QA~\cite{zhou2025egotextvqa,zhou2025scene_text_grounding}, low-light settings~\cite{zhang2025egonight}, and assistive scenarios~\cite{xiaoegoblind,wang2025streameqa,wang2026towards_explainable,wang2026dual_space_intervention,li2026collaborated_hallucination,sultan2026walkgpt,yuan2025walkvlm,zamiri2026say_conversational,xiao2026ego_grounding} further round out the evaluation landscape. Taken together, these benchmarks span a rich set of reasoning skills but do not require joint ego-exo memory, which is the gap EgoExoMem targets.

\begin{figure}
    \centering
    \includegraphics[width=\linewidth]{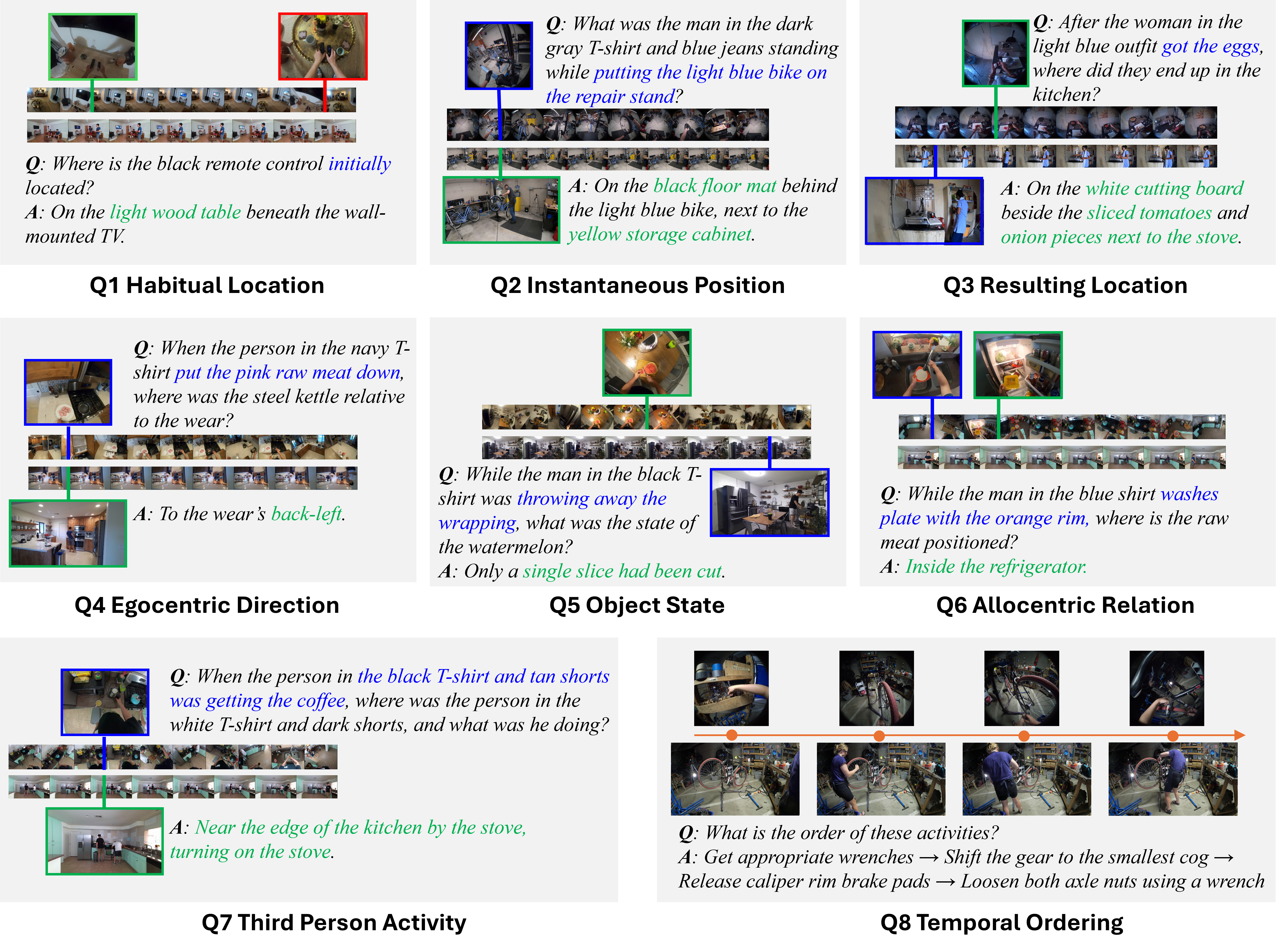}
    \caption{Illustrative examples of the eight QA types (Q1--Q8) in EgoExoMem, covering object location, spatial direction, object state, human activity, and temporal ordering. Each task requires grounding answers in visual evidence across frames, highlighting fine-grained spatial-temporal reasoning in real-world scenes.}
    \label{fig:concept}
    \vskip-4ex
\end{figure}

\section{EgoExoMem}

\subsection{Video Selection}
We select synchronized videos from two datasets, LEMMA~\cite{jia2020lemma} and EgoExo4D~\cite{grauman2024ego}. Since our goal is to capture spatial changes over time through memory, the camera wearer must interact extensively with the environment. We therefore retain only the \textit{cooking} and \textit{bike repair} scenarios from EgoExo4D, which involve substantial environmental interaction, and exclude the remaining skilled activity categories. 
To capture activities of persons other than the camera wearer, we use all multi-agent videos from LEMMA, supplemented by single-agent videos from its test set. 
The video length distributions are shown in Fig.~\ref{fig:dataset_statistics} (a): LEMMA ($M=111.3s$, $\sigma=74.8s$) and EgoExo4D ($M=425.0s$, $\sigma=339.3s$).

\subsection{QA Types}
We refer to QA types from egocentric episodic memory~\cite{grauman2022ego4d, yang2025egolife} and spatial intelligence benchmarks~\cite{yang2025thinking, chen2025savvy, majumdar2023openeqa}, adapting them to the synergistic nature of ego-exo memory as shown in Fig.~\ref{fig:concept}.

\noindent\textbf{Habitual Location (HL)} tests persistent spatial map retention by asking where an object is habitually located, requiring semantic spatial memory integrated over the full video. Ego and exo views contribute complementarily: the egocentric stream resolves fine-grained proximity, while the exocentric view anchors answers in global scene layout. \textbf{Instantaneous Position (IP)} probes the model's ability to bridge egocentric action observations and allocentric spatial positions, requiring a latent mapping between first-person and world-centered reference frames. \textbf{Resulting Location (RL)}
evaluates spatial memory across an event boundary, combining temporal grounding with spatial memory update to track an object's location after manipulation. \textbf{Egocentric Direction (ED)}
assesses fine-grained spatial awareness within the first-person reference frame, asking where a target lies relative to the camera wearer (\textit{e.g.}, back-left). While ego-view-anchored, exocentric frames provide complementary global context to resolve directions occluded from the wearer's perspective. Models must maintain a continuous spatial representation as the wearer moves. \textbf{Object State (OS)}
tests whether the model identifies the physical state of a manipulated object at a specified moment (\textit{e.g.}, uncut/sliced, whole/eaten). Object states change incrementally as procedural actions unfold, demanding frame-accurate understanding of manipulation sequences from either or both views. \textbf{Allocentric Relation (AR)} probes allocentric spatial reasoning by asking where one object lies relative to another (\textit{e.g.}, left of, directly behind). Both ego and exo views are viable but imperfect, as each provides a different vantage on the same configuration, potentially requiring cross-view integration. \textbf{Third Person Activity (TPA)}
requires simultaneous spatial tracking of two people: while one agent performs a stated action, the model must locate the second agent in the scene. It benefits substantially from the exocentric view, which captures the full room layout, including agents outside the egocentric field of view. \textbf{Temporal Ordering (TO)} evaluates episodic memory for procedural sequences by requiring models to arrange sub-actions into their correct temporal order. The primary challenge is temporal-episodic: models must recall event ordering across potentially long video segments, where the egocentric view provides richer action-level detail and the exocentric view aids in disambiguating simultaneous or spatially separated sub-events.

\subsection{Benchmark Construction}
\label{sec:benchmark_construction}
\begin{wrapfigure}{r}{0.4\textwidth}
    \centering
    \includegraphics[width=\linewidth]{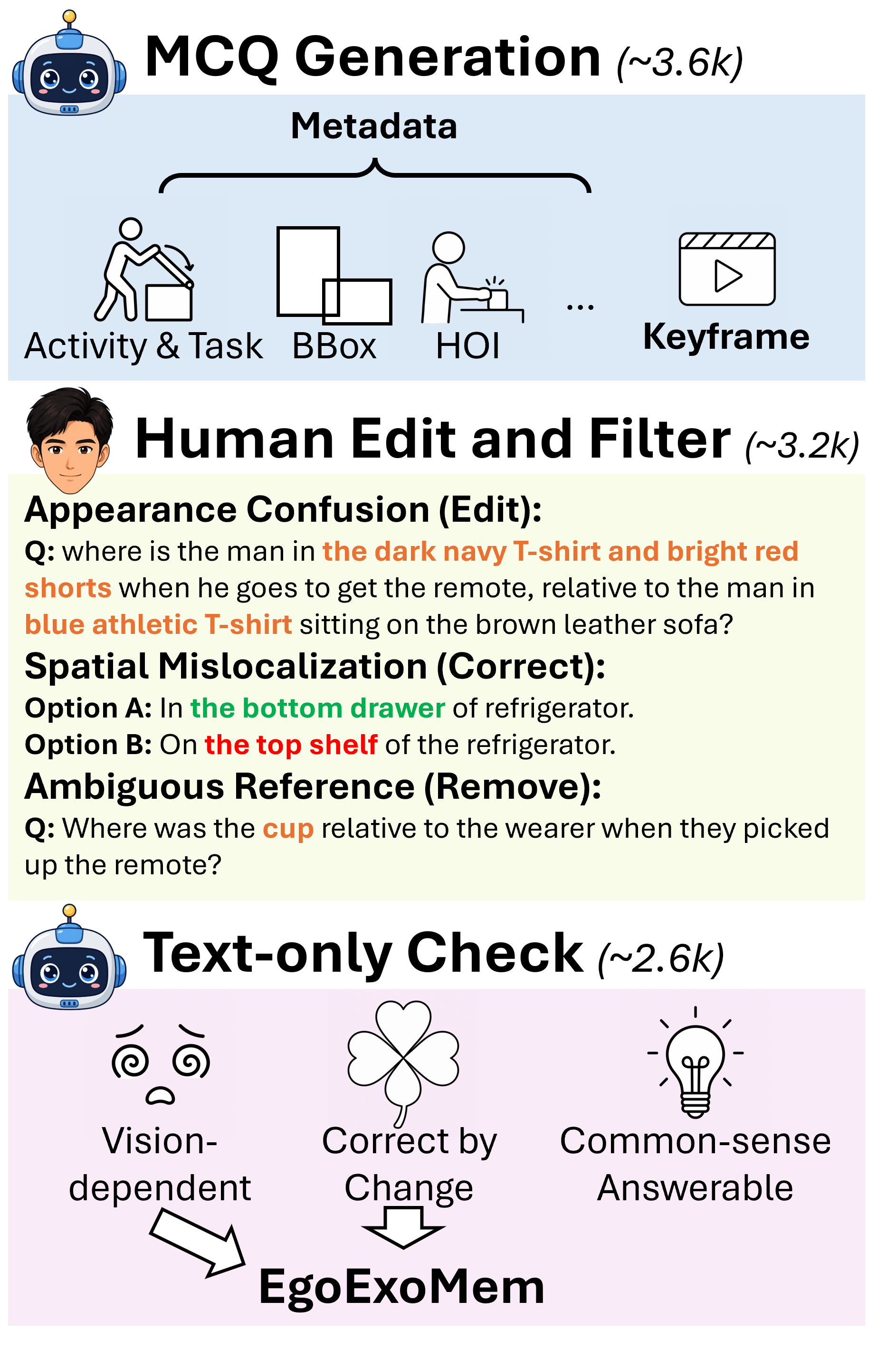}
    \caption{Overview of the EgoExoMem benchmark construction pipeline.}
    \vskip-4ex
    \label{fig:pipeline}
\end{wrapfigure}

Fig.~\ref{fig:pipeline} illustrates the benchmark construction pipeline: MCQs are first generated, then human-edited and filtered for accuracy, and finally subjected to a text-only check to ensure vision dependency.
\noindent\textbf{MCQ Generation.}
QA pairs are generated programmatically from structured annotations and then verbalized using GPT-5.4~\cite{openai2026gpt54}. The pipeline is shared across both datasets but draws on different annotation sources. 
Each question is formatted as a four-option multiple-choice question with exactly one correct answer.

For LEMMA, the primary annotation source is per-frame HOI labels, which record tuples of (action verb, interacted object, furniture anchor location) alongside person bounding boxes in the exocentric view. 
We extract contiguous action segments and derive structured ground-truth facts (object resting location, person position, object displacement, egocentric object direction, state transitions) as seeds for question generation. 
For EgoExo4D, we use keystep annotations (step name, step description, start/end timestamps) to identify activity segments, and relation annotations (per-frame object bounding box tracks) to accurately localize objects in the scene. 
Egocentric object directions are computed from the median of bounding box centroid across a track sequence.

For both datasets, each candidate is passed to GPT-5.4 with up to two keyframe images and a structured ground-truth context, producing a natural-language question and four answer options. People are identified by visual appearance rather than abstract labels. 
The keyframe selection varies by question type: \textbf{HL} uses a representative exocentric frame at the dominant get/put action; \textbf{IP} uses a window of ±5 surrounding exocentric frames around the action timestamp; \textbf{RL} uses an egocentric frame at pickup and an exocentric frame at placement; \textbf{ED} uses an egocentric frame, with the egocentric direction derived from the interacted object's bounding box centroid; \textbf{OS} uses keyframes at both the state-changing event and the later query action; \textbf{AR} uses a single exocentric keyframe, with the spatial relation between two objects determined entirely by the model from the image; \textbf{TO} is derived directly from annotation timestamps. \textbf{TPA} (simultaneous two-person localization) is generated for LEMMA only, as EgoExo4D lacks concurrent multi-person spatial annotations.

\noindent\textbf{Human Editing and Filtering.} 
The generated MCQs are not guaranteed to be fully accurate. 
To ensure dataset quality, three human annotators reviewed all generated MCQs and either corrected them (for instance, fixing confusion over human appearance, location ambiguity, or incorrect give/get relations) or discarded them entirely (such as those with ambiguous reference objects or objects that never appear in the video). The interface used for human verification is shown in Fig.~\ref{fig:verification}.

\begin{figure}
    \centering
    \includegraphics[trim=0 10 0 10, clip, width=0.85\linewidth]{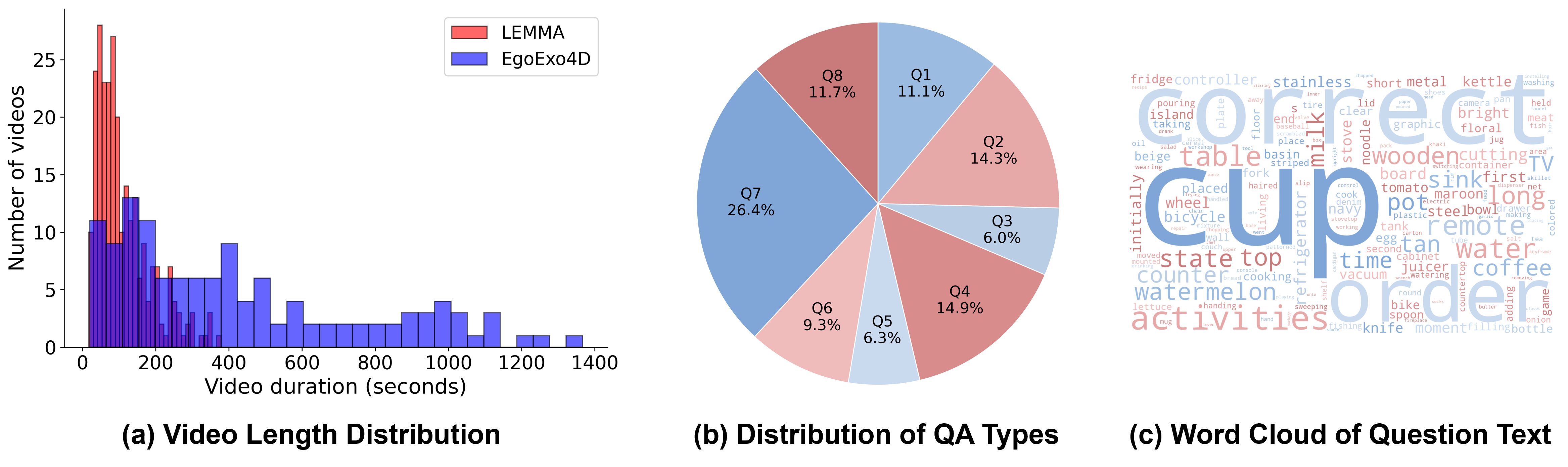}
    \caption{Dataset statistics of EgoExoMem. (a) Video length distribution for LEMMA and EgoExo4D subsets. (b) Distribution of QA types across the benchmark. (c) Word cloud of question text, illustrating the diversity of spatial and temporal vocabulary.}
    \label{fig:dataset_statistics}
    \vskip-3ex
\end{figure}

\noindent\textbf{Text-only Check.} 
The label noise in the MCQs is minimized through human editing and filtering. 
To ensure the dataset remains challenging and vision-dependent, we prompt GPT-4o~\cite{openai2024gpt4o} at temperature values of 0 and 1 to answer each MCQ based solely on the question text, without any video input. 
If GPT-4o answers correctly at both temperatures, the question is deemed common-sense answerable; if incorrect at both, vision-dependent; if mixed, potentially chance-correct. 
We discard all common-sense answerable questions and remove a subset of chance-correct ones to ensure that the text-only accuracy of GPT-4o ($Temperature=0$) for each QA type remains below $45\%$. 
The resulting distribution of QA types and the word cloud of questions are shown in Fig.~\ref{fig:dataset_statistics} (b-c).

\section{E$^2$-Select: EgoExo Frame Selection}
Unlike prior ego-exo benchmarks that focus on view matching or local clip understanding, EgoExoMem treats synchronized egocentric and exocentric videos as a persistent dual-view memory. It requires models to integrate complementary evidence across time: ego views provide fine-grained hand-object and wearer-relative cues, whereas exo views offer global layout, other-agent activity, and out-of-view context.

To address this setting, we propose E$^2$-Select, a training-free frame selection method with independent view scoring, relevance-based budget allocation, and per-view k-DPP sampling. 
This design preserves view-specific evidence, reduces temporal redundancy, and implicitly routes each query to the most informative ego-exo memory evidence.

\noindent\textbf{Independent View Scoring.}
We first conduct pre-selection of individual views according to the query-visual relevance.
We compute query-frame relevance independently for each view to avoid cross-view interference.
Given a query $\mathbf{q}$, ego frames $\{\mathbf{f}_i^e\}_{i=1}^{N_e}$, and exo frames $\{\mathbf{f}_j^x\}_{j=1}^{N_x}$ where $N_e$ denotes the number of frames of egocentric video and $N_x$ denotes the number of frames of exocentric video, we compute CLIP~\cite{radford2021learning} cosine similarities separately according to Eq.~\ref{eq:1} and Eq.~\ref{eq:2}. 
\begin{equation}
\label{eq:1}
    \mathbf{s}_e(i) = \text{sim}\!\left(\phi_{visual}(\mathbf{f}_i^e),\, \phi_{text}(\mathbf{q})\right), \quad i = \left[1, \ldots, N_e\right],
\end{equation}
\begin{equation}
\label{eq:2}
    \mathbf{s}_x(j) = \text{sim}\!\left(\phi_{visual}(\mathbf{f}_j^x),\, \phi_{text}(\mathbf{q})\right), \quad j = \left[1, \ldots, N_x\right],
\end{equation} 
where $\phi_{visual}(\cdot)$ and $\phi_{text}(\cdot)$ denote the CLIP visual encoder and text encoder. Each view is evaluated independently in its own semantic subspace, which avoids cross-view score interference and prevents a more dominant view from overwhelming the other during joint retrieval. When the two views are synchronized, $N_e$ is equal to $N_x$.

\noindent\textbf{Relevance-Based Budget Allocation.}
To address view asymmetry, we allocate the total frame budget $K$ proportionally to the aggregated query relevance of each view (Eq.~\ref{eq:3}): a view whose frames are collectively more relevant to the query receives a larger frame budget, providing a soft and query-dependent alternative to hard view selection.
\begin{equation}
\label{eq:3}
    K_e = \left\lfloor K \cdot \frac{\sum_{i} \mathbf{s}_e(i)}{\sum_{i} \mathbf{s}_e(i) + \sum_{j} \mathbf{s}_x(j)} \right\rceil, \quad K_x = K - K_e,
\end{equation}
\noindent where $\lfloor \cdot \rceil$ denotes rounding to the nearest integer. This formulation implicitly performs view selection while maintaining a better balance than hard-selecting between ego and exo views at each timestamp.

\noindent\textbf{Per-View k-DPP Sampling.}
To address temporal redundancy within each view, we replace coverage-based selection with a Determinantal Point Process (DPP), which provides a principled diversity guarantee via determinant geometry.

For the ego view, we construct a quality-diversity kernel matrix $\mathbf{L}^e \in \mathbb{R}^{N_e \times N_e}$, as shown in Eq.~\ref{eq:4}.
\begin{equation}
\label{eq:4}
    \mathbf{L}^e_{ij} = \mathbf{s}_e(i) \cdot \text{sim}\!\left(\phi_{visual}(\mathbf{f}_i^e),\, \phi_{visual}(\mathbf{f}_j^e)\right) \cdot \mathbf{s}_e(j),
\end{equation}
\noindent The diagonal entries $\mathbf{L}^e_{ii} = \mathbf{s}_e(i)^2$ encode per-frame relevance; off-diagonal entries penalize the simultaneous selection of pairs of frames with similar visual content.
We then draw a subset $S_e$ of size $K_e$ from the k-DPP defined by $\mathbf{L}^e$ as in Eq.~\ref{eq:5}.
\begin{equation}
\label{eq:5}
    P(S_e) \propto \det\!\left(\mathbf{L}^e_{S_e}\right), \quad |S_e| = K_e,
\end{equation}
\noindent 
where $\mathbf{L}^e_{S_e}$ denotes the principal submatrix of $\mathbf{L}^e$ restricted to the frame indices in $S_e$.
Because the determinant measures the volume spanned by the selected frame embeddings, maximizing it encourages a subset that is both relevant to the query and visually diverse, rather than merely well-spaced in time.

We apply the same k-DPP sampling procedure to the exocentric stream, obtaining a subset $S_x$ with $|S_x|=K_x$.
For efficiency, we use approximate Cholesky-based k-DPP~\cite{derezinski2019exact} inference, which has complexity $\mathcal{O}(N^2K)$.

\noindent\textbf{Timestamp-Ordered Merge.}
After sampling view-specific subsets, we merge the selected ego and exo frames according to their original timestamps as shown in Eq.~\ref{eq:6}.
\begin{equation}
\label{eq:6}
    \mathcal{F}
    =
    \operatorname{Sort}_{t}\!\left(
    \{\mathbf{f}_i^e \mid i \in S_e\}
    \cup
    \{\mathbf{f}_j^x \mid j \in S_x\}
    \right),
\end{equation}
where $t$ denotes the original video timestamps. This chronological merge preserves the temporal structure of the synchronized ego-exo memory while retaining view-specific evidence. The resulting sequence $\mathcal{F}$ contains $K$ frames and is used as the visual input to the MLLM.

\section{Experiments}
\label{sec:experiments}
Since we propose the first benchmark dedicated to egocentric and exocentric memory, EgoExoMem, we dedicate ourselves to answering four research questions: (1) \textit{Can egocentric and exocentric streams serve as complementary memory sources for spatial and temporal reasoning?} (2) \textit{How do existing MLLMs perform on EgoExoMem?} (3) \textit{How does E$^2$-Select compare to existing memory retrieval strategies on EgoExoMem?} (4) \textit{under what conditions does joint ego-exo memory retrieval deteriorate, and what are the underlying causes?}
\begin{table}[t]
\centering
\caption{Per-category accuracy (\%) of MLLMs on EgoExoMem across different input views.}
\resizebox{\linewidth}{!}{%
\begin{tabular}{llllccccccccr}
\toprule
\textbf{Model}& \textbf{Frame Input}&\textbf{Views}&  \textbf{HL} & \textbf{IP} & \textbf{RL} & \textbf{ED} & \textbf{OS} & \textbf{AR} & \textbf{TPA} & \textbf{TO} & \textbf{Avg}\\
\midrule
Gemini 2.5 Flash~\cite{comanici2025gemini}&concat&Ego&58.1&62.5&63.5&\textbf{38.9}&\textbf{56.4}&54.4&\textbf{48.9}&34.7&52.2\\
Gemini 2.5 Flash~\cite{comanici2025gemini}&concat&Exo&51.4 & \textbf{72.2} & 59.6 & 34.6 & 48.8 & 44.6 & 45.5 & 38.9 & 49.5\\
\rowcolor{gray!20}
Gemini 2.5 Flash~\cite{comanici2025gemini}&concat&Ego+Exo&\textbf{62.9} & 70.9 & \textbf{66.7} & 38.7 & 53.7 & \textbf{56.3} & 48.0 & \textbf{44.9} & \textbf{55.3}\\
\midrule
InternVL3.5~\cite{wang2025internvl3} &concat&Ego                  & 50.7 & 55.5 & 55.1 & \textbf{38.2} & 55.5 & \underline{43.3} & \underline{47.7} & 39.3 & 48.2 \\
InternVL3.5~\cite{wang2025internvl3} &interleaved &Ego      & \textbf{53.8} & 57.7 & 57.7 & 36.1 & \underline{56.7} & 42.5 & \textbf{49.5} & \underline{41.9} & \underline{49.5} \\
InternVL3.5~\cite{wang2025internvl3} &concat&Exo                  & 47.2 & \underline{65.4} & 50.6 & 33.0 & 50.0 & 39.2 & 36.0 & 37.0 & 44.8 \\
InternVL3.5~\cite{wang2025internvl3} &interleaved &Exo      & 46.9 & \textbf{66.3} & 51.3 & 34.0 & 44.5 & 37.9 & 35.7 & 38.0 & 44.3 \\
\rowcolor{gray!20}
InternVL3.5~\cite{wang2025internvl3} &concat&Ego+Exo                & \underline{53.1} & 65.2 & \textbf{64.1} & \underline{36.4} & \textbf{57.9} & 42.1 & 44.8 & \underline{41.9} & \textbf{50.7} \\
\rowcolor{gray!20}
InternVL3.5~\cite{wang2025internvl3} &interleaved &Ego+Exo  & 50.7 & 63.1 & \underline{58.3} & 35.3 & 55.5 & \textbf{44.2} & 39.2 & \textbf{42.2} & 48.6 \\
\midrule
LLaVA-OV~\cite{li2024llava} &concat&Ego                     & \underline{48.3} & 53.6 & 52.6 & 32.7 & 50.0 & 37.9 & \underline{36.9} & \underline{36.3} & 43.5 \\
LLaVA-OV~\cite{li2024llava} &interleaved &Ego         & \textbf{48.6} & 55.3 & \textbf{56.4} & \textbf{35.6} & 51.8 & 38.3 & \textbf{37.0} & 35.3 & 44.8 \\
LLaVA-OV~\cite{li2024llava} &concat&Exo                     & 47.2 & \textbf{64.2} & 42.9 & 31.2 & 50.0 & 37.9 & 34.1 & \textbf{36.6} & 43.0 \\
LLaVA-OV~\cite{li2024llava} &interleaved &Exo         & 46.2 & 62.5 & 42.3 & 29.6 & 50.6 & 36.7 & 34.8 & 35.0 & 42.2 \\
\rowcolor{gray!20}
LLaVA-OV~\cite{li2024llava} &concat&Ego+Exo                   & 51.0 & \underline{63.6} & 50.0 & 34.8 & \underline{52.4} & \textbf{41.2} & 34.6 & 35.6 & \textbf{45.4} \\
\rowcolor{gray!20}
LLaVA-OV~\cite{li2024llava} &interleaved &Ego+Exo     & 47.6 & 60.1 & \underline{53.2} & \underline{35.3} & \textbf{53.7} & \textbf{41.2} & 33.1 & 36.0 & \underline{45.0} \\
\midrule
Qwen2.5-VL~\cite{bai2025qwen25vl} &concat&Ego                   & 50.0 & 49.3 & 53.2 & \textbf{36.9} & \underline{43.3} & 38.3 & \textbf{38.5} & \textbf{37.0} & 43.3 \\
Qwen2.5-VL~\cite{bai2025qwen25vl} &interleaved &Ego       & 51.7 & 50.4 & \underline{57.1} & \underline{34.0} & 40.9 & 38.3 & 35.3 & 35.0 & 42.8 \\
Qwen2.5-VL~\cite{bai2025qwen25vl} &concat&Exo                   & 50.7 & 57.1 & 51.9 & 30.4 & 41.5 & 40.0 & \underline{35.7} & 36.6 & 43.0 \\
Qwen2.5-VL~\cite{bai2025qwen25vl} &interleaved &Exo       & 50.0 & 56.1 & 53.2 & 30.1 & 37.8 & \underline{40.4} & 35.6 & \textbf{37.0} & 42.5 \\
\rowcolor{gray!20}
Qwen2.5-VL~\cite{bai2025qwen25vl} &concat&Ego+Exo                 & \underline{53.1} & \textbf{58.2} & \textbf{58.3} & 31.7 & 42.7 & \underline{40.4} & 34.3 & 36.0 & \underline{44.3} \\
\rowcolor{gray!20}
Qwen2.5-VL~\cite{bai2025qwen25vl} &interleaved &Ego+Exo   & \textbf{53.5} & \underline{57.4} & 54.5 & \underline{34.0} & \textbf{45.7} & \textbf{43.3} & 32.9 & 36.0 & \textbf{44.7} \\

\bottomrule
\end{tabular}%
}
\label{tab:all_frame_concat}
\vskip-2ex
\end{table}
\begin{table}[t]
\caption{Per-category accuracy (\%) of different memory mechanisms on EgoExoMem.}
\centering
\resizebox{\linewidth}{!}{%
\begin{tabular}{llccccccccr}
\toprule
\textbf{Method} & \textbf{Views}&  \textbf{HL} & \textbf{IP} & \textbf{RL} & \textbf{ED} & \textbf{OS} & \textbf{AR} & \textbf{TPA} & \textbf{TO} & \textbf{Avg}\\
\midrule
AKS~\cite{tang2025adaptive}&Ego&\textbf{64.7}&64.2&62.8&37.7&56.1&47.9&\textbf{55.9}&44.8&54.2\\
AKS~\cite{tang2025adaptive}&Exo&50.0&68.5&54.5&33.3&46.3&42.9&42.2&38.3&47.0\\
BOLT~\cite{liu2025bolt}&Ego&\underline{62.9}&63.6&62.2&\textbf{40.3}&58.5&\underline{51.3}&\underline{55.6}&\underline{45.9}&\underline{55.0}\\
BOLT~\cite{liu2025bolt}&Exo&51.0&69.5&59.6&34.3&44.5&43.8&41.7&39.6&48.0\\
BM25~\cite{robertson2009bm25}& Ego   & 51.7 & 59.6 & 62.8 & 34.3 & 51.8 & 44.6 & 46.0 & 45.2 & 49.5 \\
BM25~\cite{robertson2009bm25}& Exo   & 47.9 & 66.0 & 57.7 & 30.6 & 43.9 & 44.2 & 39.8 & 37.6 & 46.0 \\
  \rowcolor{gray!20}
BM25~\cite{robertson2009bm25} & Ego+Exo & 54.2 & 66.3 & 62.8 & 37.4 & 51.8 & 45.8 & 43.8 & 42.9 & 50.6 \\
DPR~\cite{karpukhin2020dense} & Ego   & 52.8 & 60.6 & 64.7 & 34.5 & 48.2 & 44.2 & 45.8 & 44.9 & 49.5 \\
DPR~\cite{karpukhin2020dense} & Exo   & 48.6 & 66.3 & 54.5 & 29.1 & 42.1 & 41.2 & 39.7 & 37.6 & 44.9 \\
  \rowcolor{gray!20}
DPR~\cite{karpukhin2020dense} & Ego+Exo & 56.3 & 68.2 & \textbf{67.3} & 36.9 & 45.7 & 40.8 & 45.8 & 41.9 & 50.4 \\
VideoRAG~\cite{jeong2025videorag} & Ego& 47.2 & 53.6 & 62.8 & 36.6 & 53.0 & 40.8 & 48.0 & 40.3 & 47.8\\ 
VideoRAG~\cite{jeong2025videorag} & Exo   & 47.2 & 69.0 & 60.3 & 29.6 & 46.3 & 37.9 & 41.4 & 36.0 & 46.0 \\
  \rowcolor{gray!20}
VideoRAG~\cite{jeong2025videorag}& Ego+Exo & 49.0 & 66.6 & 62.2 & 33.5 & \textbf{57.9} & 43.8 & 41.0 & 39.6 & 49.2 \\
Video-RAG~\cite{luo2024video}&Ego        & 56.3 & 55.3 & 50.0 & 36.9 & 50.6 & 47.1 & 48.8 & 35.6 & 47.6 \\
Video-RAG~\cite{luo2024video} &Exo        & 49.3 & 63.9 & 51.9 & 34.0 & 42.7 & 43.8 & 40.6 & 35.3 & 45.2 \\
\rowcolor{gray!20}
Video-RAG~\cite{luo2024video} &Ego+Exo      & 53.1 & 66.0 & 57.1 & 35.8 & 50.6 & 45.8 & 43.0 & 35.0 & 48.3 \\
\rowcolor{gray!20}
EgoMAS~\cite{kim2026ma} &Ego+Exo      & 45.5 & 56.6 & 51.3 & 28.6 & 62.8 & 35.8 & 37.0 & 38.3 & 44.5 \\
\rowcolor{gray!20}
WorldMM~\cite{yeo2025worldmm}&Ego+Exo&36.5&44.8&39.4&23.1&36.5&36.3&35.1&31.5&35.4\\
\rowcolor{gray!40}
\midrule
E$^2$-Select (ours)&Ego+Exo&59.8&\textbf{72.8}&\underline{64.1}&34.3&\textbf{57.9}&\textbf{51.7}&47.9&\textbf{46.2}&\textbf{56.3}\\ 
\bottomrule
\end{tabular}%
}
\label{tab:baselines}
\vskip-2ex
\end{table}
\begin{table}[]
\centering
\caption{Ablation study of E$^2$-Select.}
\resizebox{\linewidth}{!}{%
\begin{tabular}{llllccccccccr}
\toprule
\textbf{Model}& \textbf{Frame Input}&\textbf{Views}& \textbf{HL} & \textbf{IP} & \textbf{RL} & \textbf{ED} & \textbf{OS} & \textbf{AR} & \textbf{TPA} & \textbf{TO} & \textbf{Avg}\\
\midrule
         Gemini 2.5 Flash~\cite{comanici2025gemini}&concat&Ego+Exo&62.9 & 70.9 & 66.7 & \textbf{38.7} & 53.7 & \textbf{56.3} & 48.0 & 44.9 & 55.3\\
         Gemini 2.5 Flash~\cite{comanici2025gemini}&k-DPP&Ego& 67.8&69.5&68.6&38.4&57.9&\textbf{56.3}&48.6&45.9&56.6\\
         Gemini 2.5 Flash~\cite{comanici2025gemini}&k-DPP+hard selection&Ego+Exo& 67.1&73.0&\textbf{69.0}&34.1&57.3&54.4&\textbf{52.3}&46.7&56.7\\
         \rowcolor{gray!20}
         Gemini 2.5 Flash~\cite{comanici2025gemini}&k-DPP+soft allocation~\textbf{(E$^2$-Select)}&Ego+Exo&\textbf{72.5}&\textbf{77.0}&66.0&36.8&\textbf{60.4}&54.2&51.6&\textbf{47.4}&\textbf{58.2}\\
         \midrule
         InternVL3.5~\cite{wang2025internvl3} &concat&Ego+Exo                & 53.1 & 65.2 & \textbf{64.1} & 36.4 & \textbf{57.9} & 42.1 & 44.8 & 41.9 & 50.7 \\
         InternVL3.5~\cite{wang2025internvl3}&k-DPP&Ego& \textbf{63.3}&63.3&62.2&35.3&54.9&52.1&\textbf{55.1}&44.6&53.8\\
         InternVL3.5~\cite{wang2025internvl3}&AKS~\cite{tang2025adaptive}+soft allocation&Ego+Exo&58.0 & 71.7 & 68.6 & 37.1 & 58.5 & 46.2 & 45.8 & 42.6 & 53.6\\
         InternVL3.5~\cite{wang2025internvl3}&BOLT~\cite{liu2025bolt}+soft allocation&Ego+Exo&56.6 & 68.2 & 64.1 & 37.9 & 54.3 & 47.9 & 44.1 & 43.6 & 52.1\\
         InternVL3.5~\cite{wang2025internvl3}&k-DPP+hard selection&Ego+Exo&60.5&71.7&63.5&\textbf{38.2}&52.4&\textbf{53.8}&45.1&43.6&53.6\\
         \rowcolor{gray!20}
         InternVL3.5~\cite{wang2025internvl3}&k-DPP+soft allocation~\textbf{(E$^2$-Select)}&Ego+Exo&59.8&\textbf{72.8}&\textbf{64.1}&34.3&\textbf{57.9}&51.7&47.9&\textbf{46.2}&\textbf{56.3}\\
         \bottomrule
    \end{tabular}}
    \label{tab:ablation}
    \vskip-2ex
\end{table}
\subsection{Implementation Details}
\label{sec:implementation}
We evaluate a proprietary MLLM, Gemini 2.5 Flash~\cite{comanici2025gemini}, and open-source ones, including InternVL3.5~\cite{wang2025internvl3}, LLaVA-OneVision~\cite{li2024llava}, and Qwen2.5-VL~\cite{bai2025qwen3}, to investigate the synergy of egocentric and exocentric memory and their performance on EgoExoMem. 
For this part, we explore two strategies for inputting frames from two video sources: concatenation~\cite{kim2026ma, he2025egoexobench}, which preserves intra-video temporal consistency but provides weaker inter-video frame correspondence; and interleaving~\cite{li2024llava_next_interleave}, which improves inter-video frame correspondence but disrupts intra-video temporal consistency. 
Based on the best-performing open-source MLLM, InternVL3.5, we further experiment with different memory mechanisms, RAG-based~\cite{robertson2009bm25, karpukhin2020dense, luo2024video, jeong2025videorag} and frame selection for single video~\cite{tang2025adaptive, liu2025bolt}, and our E$^2$-Select for two synchronized videos. 

For fair evaluation, all reasoning models receive $32$ frames as input. For MLLMs, $16$ frames are uniformly sampled from either or both videos. For memory mechanisms, all videos are first processed at $1$ FPS. For the RAG-based methods, videos are segmented into clips, and the top-$k$ ($k=8$) most relevant clips are retrieved, each represented by $4$ frames. The captions used for retrieval are generated with Gemini 2.5 Flash~\cite{comanici2025gemini}. 
All other settings follow the original configurations of the methods. Our experiments are conducted on 4 NVIDIA A100 (40GB) GPUs.



\subsection{Results}
\noindent\textbf{Reasonability of EgoExoMem and Performance of MLLMs.}
Tab.~\ref{tab:all_frame_concat} evaluates MLLMs with two simple video frame input strategies: concatenation and interleaving. 
To assess the complementarity of the two views, we keep the total number of frames constant by duplicating the same video for single-video input, ensuring any performance gain stems from view complementarity rather than additional temporal information. 
Overall, combining both views consistently outperforms single-view input across all evaluated models. 
\textit{This supports the reasonability of leveraging both views in EgoExoMem.} 
For instance, Gemini 2.5 Flash improves from $52.2\%$ (Ego) and $51.4\%$ (Exo) to $55.3\%$ with Ego+Exo, and InternVL3.5 improves from $48.2\%$ (Ego) and $44.3\%$ (Exo) to $50.7\%$ with Ego+Exo under concatenation. For Habitual Location, Resulting Location, Object State, and Allocentric Relation, the cross-view combination is clearly preferred across models. 
Instantaneous Position favors the exocentric view, and Egocentric Direction favors the egocentric view. Notably, Third Person Activity consistently favors the egocentric view across all models, with InternVL3.5 showing a gap as large as $13.8\%$ ($49.5\%$ \textit{vs.} $35.7\%$), which is counterintuitive and will be analyzed in the following section. Temporal Ordering shows no significant difference across egocentric, exocentric, and mixed inputs, with scores remaining comparable across all input strategies.

Despite being state-of-the-art MLLMs, all models achieve relatively low average scores, with the best-performing model, Gemini 2.5 Flash, reaching only $55.3\%$, highlighting the challenge of EgoExoMem. Among open-source models, InternVL3.5 achieves the best performance and is therefore adopted as the reasoning model for the subsequent memory mechanism evaluation.

\noindent\textbf{Performance of Memory Mechanisms.} Tab.~\ref{tab:baselines} demonstrates the performance of memory mechanisms in order of increasing complexity, including frame selection (AKS~\cite{tang2025adaptive} and BOLT~\cite{liu2025bolt}), text retrieval (BM25~\cite{robertson2009bm25} and DPR~\cite{karpukhin2020dense}), visually-grounded retrieval (Video-RAG~\cite{luo2024video} and VideoRAG~\cite{jeong2025videorag}), and structured memory retrieval (EgoMAS~\cite{kim2026ma} and WorldMM~\cite{yeo2025worldmm}). Consistent with the frame combination results in Tab.~\ref{tab:all_frame_concat}, retrieving from both egocentric and exocentric streams yields better performance than single-view inputs across RAG-based methods. However, despite their strong performance on long-video benchmarks, structured retrieval methods underperform on EgoExoMem, whereas simpler approaches, frame selection and text retrieval, prove more effective. We attribute this to the minute-level duration of our videos: for such short clips, complex retrieval pipelines introduce unnecessary overhead and may be redundant~\cite{xue2025adavideorag}.

As frame selection methods demonstrate superior performance on single-view understanding, we propose E$^2$-Select, the first frame selection method for dual-view ego-exo inputs. 
As shown in Tab.~\ref{tab:baselines}, E$^2$-Select achieves superior performance over all baselines. 
A comprehensive ablation study is further conducted to verify the design choices in Tab.~\ref{tab:ablation}. 
Though AKS~\cite{tang2025adaptive} and BOLT~\cite{liu2025bolt} achieve better performance on single-view ego input, k-DPP~\cite{kulesza2011kdpps} combines the two views more effectively. 
As AKS and BOLT rely on temporal structure estimation and single-video saliency calibration, their performance degrades under cross-view budget allocation. 
In contrast, k-DPP natively supports any frame budget $k$ and selects frames by maximizing diversity in feature space, making it distribution-agnostic and robust to the domain shift between ego and exo views. 
We also compare soft allocation with hard selection, which picks the more query-similar view at each timestep and then applies k-DPP.
It performs on par with single-view k-DPP, failing to fully exploit the complementarity of dual views.

\begin{figure}
    \centering
    \includegraphics[trim=0 10 0 10, clip, width=\linewidth]{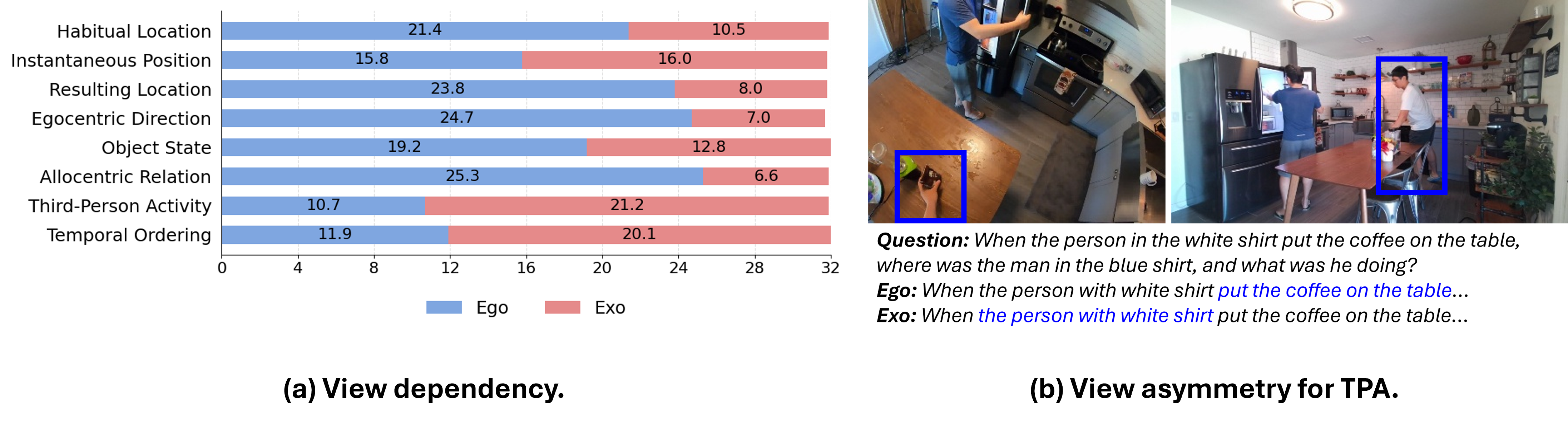}
    \caption{Failure case analysis. (a) Question-aware view dependency measured by CLIP similarity across all QA types. (b) View-specific emphasis on different cues, such as action in the ego view and appearance in the exo view, even when the answers are visible in both views.}
    \label{fig:failure}
    \vskip-2ex
\end{figure}

\noindent\textbf{Failure Cases and Potential Reasons.} Among all baselines, it is counterintuitive that Third Person Activity relies significantly on the egocentric view. 
By examining the keyframes used to generate the MCQs, we find that many questions can be answered from either view due to the small room settings in LEMMA~\cite{jia2020lemma} and the collaborative nature of the tasks, as shown in Fig.~\ref{fig:failure} (b). 
This raises the question of \textit{whether the view most relevant to the answer is also the most relevant to the question}.
To investigate this, we report the question-aware view-dependency measured by CLIP for all question types in Fig.~\ref{fig:failure} (a). 
We observe that the view preferences of the question and the answer for Third Person Activity (Tab.~\ref{tab:all_frame_concat} and Tab.~\ref{tab:baselines}) are significantly different: the question favors the exocentric view, whereas the answer favors the egocentric view. This discrepancy causes severe degradation of TPA performance when frame selection is based solely on question-aware similarity, highlighting the necessity of synergy between both views.

\section{Conclusion}
We present EgoExoMem, the first benchmark for memory-based reasoning over synchronized ego-exo video. Spanning eight QA types across spatial, temporal, and cross-view
memory, it reveals that neither view alone suffices for comprehensive understanding, and that existing MLLMs and memory mechanisms fail to fully exploit dual-view complementarity. To fill the gap in multi-view frame selection, we propose E$^2$-Select, which achieves superior performance via relevance-based budget allocation and k-DPP sampling that accounts for view asymmetry and cross-view temporal consistency. Failure analysis further exposes a systematic view-dependency mismatch for Third Person Activity, motivating joint query-answer view routing in future work. We hope EgoExoMem and E$^2$-Select serve as a foundation for cross-view memory reasoning in embodied AI.

\section*{Acknowledgment}
 This work was performed on the HoreKa supercomputer funded by the Ministry of Science, Research and the Arts Baden-Württemberg and by the Federal Ministry of Education and Research. The authors also acknowledge support by the state of Baden-Württemberg through bwHPC and the German Research Foundation (DFG) through grant INST 35/1597-1 FUGG. The project is funded by the Deutsche Forschungsgemeinschaft (DFG, German Research Foundation) – SFB 1574 – 471687386.
This project is also supported in part by the National Natural Science Foundation of China under Grant No. 62473139, in part by the Hunan Provincial Research and Development Project (Grant No. 2025QK3019), and in part by the State Key Laboratory of Autonomous Intelligent Unmanned Systems (the opening project number ZZKF2025-2-10).
This research was partially funded by the Ministry of Education and Science of Bulgaria (support for INSAIT, part of the Bulgarian National Roadmap for Research Infrastructure).

\clearpage
\bibliographystyle{plain}
\bibliography{bib.bib}

@inproceedings{ginting2025enter_mind_palace,
  title={Enter the mind palace: Reasoning and planning for long-term active embodied question answering},
  author={Muhammad Fadhil Ginting and
                  Dong{-}Ki Kim and
                  Xiangyun Meng and
                  Andrzej Reinke and
                  Bandi Jai Krishna and
                  Navid Kayhani and
                  Oriana Peltzer and
                  David D. Fan and
                  Amirreza Shaban and
                  Sung{-}Kyun Kim and
                  Mykel J. Kochenderfer and
                  Ali{-}akbar Agha{-}mohammadi and
                  Shayegan Omidshafiei},
  booktitle={CoRL},
  year={2026}
}

@article{he2026bridging,
  title={Bridging perspectives: A survey on cross-view collaborative intelligence with egocentric-exocentric vision},
  author={He, Yuping and Huang, Yifei and Chen, Guo and Lu, Lidong and Pei, Baoqi and Xu, Jilan and Lu, Tong and Sato, Yoichi},
  journal={International Journal on Computer Vision},
  year={2026}
}

@inproceedings{deichler2025look_tell,
  title={Look and Tell: A Dataset for Multimodal Grounding Across Egocentric and Exocentric Views},
  author={Deichler, Anna and Beskow, Jonas},
  booktitle={NeurIPS 2025 Workshop on Space in Vision, Language, and Embodied AI},
  year={2025}
}

@inproceedings{feng2025object_shot,
  title={Object-Shot Enhanced Grounding Network for Egocentric Video},
  author={Feng, Yisen and Zhang, Haoyu and Liu, Meng and Guan, Weili and Nie, Liqiang},
  booktitle={CVPR},
  year={2025}
}

@inproceedings{xu2024retrieval_augmented_egocentric,
  title={Retrieval-Augmented Egocentric Video Captioning},
  author={Xu, Jilan and Huang, Yifei and Hou, Junlin and Chen, Guo and Zhang, Yuejie and Feng, Rui and Xie, Weidi},
  booktitle={CVPR},
  year={2024}
}

@inproceedings{liang2025fine_grained_spatiotemporal,
  title={Fine-grained spatiotemporal grounding on egocentric videos},
  author={Liang, Shuo and Zhong, Yiwu and Hu, Zi-Yuan and Tao, Yeyao and Wang, Liwei},
  booktitle={ICCV},
  year={2025}
}

@inproceedings{luo2024put_lifting,
  title={Put myself in your shoes: Lifting the egocentric perspective from exocentric videos},
  author={Luo, Mi and Xue, Zihui and Dimakis, Alex and Grauman, Kristen},
  booktitle={ECCV},
  year={2024}
}

@article{luo2024grounded_affordance,
  title={Grounded Affordance from Exocentric View},
  author={Luo, Hongchen and Zhai, Wei and Zhang, Jing and Cao, Yang and Tao, Dacheng},
  journal={International Journal of Computer Vision},
  year={2024},
  publisher={Springer US New York}
}

@inproceedings{huang2025sound_bridge,
  title={Sound bridge: Associating egocentric and exocentric videos via audio cues},
  author={Huang, Sihong and Wu, Jiaxin and Wei, Xiaoyong and Cai, Yi and Jiang, Dongmei and Wang, Yaowei},
  booktitle={CVPR},
  year={2025}
}

@inproceedings{jia2024audio_visual,
  title={The Audio-Visual Conversational Graph: From an Egocentric-Exocentric Perspective},
  author={Jia, Wenqi and Liu, Miao and Jiang, Hao and Ananthabhotla, Ishwarya and Rehg, James M. and Ithapu, Vamsi Krishna and Gao, Ruohan},
  booktitle={CVPR},
  year={2024}
}

@article{kang2023video_captioning,
  title={Video captioning based on both egocentric and exocentric views of robot vision for human-robot interaction},
  author={Kang, Soo-Han and Han, Ji-Hyeong},
  journal={International Journal of Social Robotics},
  year={2023},
  publisher={Springer}
}

@inproceedings{fu2025objectrelator,
  title={{ObjectRelator:} {Enabling} Cross-View Object Relation Understanding Across Ego-Centric and Exo-Centric Perspectives},
  author={Fu, Yuqian and Wang, Runze and Ren, Bin and Sun, Guolei and Gong, Biao and Fu, Yanwei and Paudel, Danda Pani and Huang, Xuanjing and Van Gool, Luc},
  booktitle={ICCV},
  year={2025}
}

@article{mahdi2025exo2egosyn,
  title={{Exo2EgoSyn:} {Unlocking} Foundation Video Generation Models for Exocentric-to-Egocentric Video Synthesis},
  author={Mahdi, Mohammad and Fu, Yuqian and Savov, Nedko and Pan, Jiancheng and Paudel, Danda Pani and Van Gool, Luc},
  journal={arXiv preprint arXiv:2511.20186},
  year={2025}
}

@inproceedings{ohkawa2025exo2egodvc,
  title={{Exo2EgoDVC:} {Dense} Video Captioning of Egocentric Procedural Activities Using Web Instructional Videos},
  author={Ohkawa, Takehiko and Yagi, Takuma and Nishimura, Taichi and Furuta, Ryosuke and Hashimoto, Atsushi and Ushiku, Yoshitaka and Sato, Yoichi},
  booktitle={WACV},
  year={2025}
}

@inproceedings{fan2025prvql,
  title={{PRVQL:} {Progressive} Knowledge-guided Refinement for Robust Egocentric Visual Query Localization},
  author={Bing Fan and Yunhe Feng and Yapeng Tian and James Chenhao Liang and Yuewei Lin and Yan Huang and Heng Fan},
  booktitle={ICCV},
  year={2025}
}

@inproceedings{barmann2022did_episodic,
  title={Where did {I} leave my keys?—{Episodic-memory-based} Question Answering on Egocentric Videos},
  author={B{\"a}rmann, Leonard and Waibel, Alex},
  booktitle={CVPRW},
  year={2022}
}

@inproceedings{datta2022episodic_memory,
  title={Episodic Memory Question Answering},
  author={Datta, Samyak and Dharur, Sameer and Cartillier, Vincent and Desai, Ruta and Khanna, Mukul and Batra, Dhruv and Parikh, Devi},
  booktitle={CVPR},
  year={2022}
}

@inproceedings{pan2026v2sam,
  title={V\({}^{\mbox{2}}\)-{SAM}: Marrying {SAM2} with Multi-Prompt Experts for Cross-View Object Correspondence},
  author={Pan, Jiancheng and Wang, Runze and Qian, Tianwen and Mahdi, Mohammad and Fu, Yanwei and Xue, Xiangyang and Huang, Xiaomeng and Van Gool, Luc and Paudel, Danda Pani and Fu, Yuqian},
  booktitle={CVPR},
  year={2026}
}

@article{wang2025streameqa,
  title={{StreamEQA:} {Towards} Streaming Video Understanding for Embodied Scenarios},
  author={Wang, Yifei and Li, Zhenkai and Qian, Tianwen and Zheng, Huanran and Wang, Zheng and Fu, Yuqian and Wang, Xiaoling},
  journal={arXiv preprint arXiv:2512.04451},
  year={2025}
}

@article{kulkarni2025egovita,
  title={{EgoVITA:} {Learning} to Plan and Verify for Egocentric Video Reasoning},
  author={Kulkarni, Yogesh and Fazli, Pooyan},
  journal={arXiv preprint arXiv:2511.18242},
  year={2025}
}

@inproceedings{su2026regionaligner,
  title={{RegionAligner:} {Bridging} Ego-Exo Views for Object Correspondence via Unified Text-Visual Learning},
  author={Su, Yuhao and Elhamifar, Ehsan},
  booktitle={WACV},
  year={2026}
}

@inproceedings{shi2025unsupervised_egoexo,
  title={Unsupervised Ego-and Exo-centric Dense Procedural Activity Captioning via Gaze Consensus Adaptation},
  author={Shi, Zhaofeng and Qiu, Heqian and Wang, Lanxiao and Wu, Qingbo and Meng, Fanman and Li, Hongliang},
  booktitle={AAAI},
  year={2025}
}

@inproceedings{li2026sava,
  title={{SAVA-X:} {Ego-to-exo} Imitation Error Detection via Scene-Adaptive View Alignment and Bidirectional Cross View Fusion},
  author={Li, Xiang and Qiu, Heqian and Wang, Lanxiao and Qiu, Benliu and Meng, Fanman and Xu, Linfeng and Li, Hongliang},
  booktitle={CVPR},
  year={2026}
}

@inproceedings{bonnetto2025epfl_smart_kitchen,
  title={{EPFL-Smart-Kitchen:} {An} Ego-Exo Multi-Modal Dataset for Challenging Action and Motion Understanding in Video-Language Models},
  author={Bonnetto, Andy and Qi, Haozhe and Leong, Franklin and Tashkovska, Matea and Rad, Mahdi and Shokur, Solaiman and Hummel, Friedhelm and Micera, Silvestro and Pollefeys, Marc and Mathis, Alexander},
  booktitle={NeurIPS},
  year={2025}
}

@inproceedings{xu2025egoexo_gen,
  title={{EgoExo-Gen:} {Ego-centric} Video Prediction by Watching Exo-centric Videos},
  author={Xu, Jilan and Huang, Yifei and Pei, Baoqi and Hou, Junlin and Li, Qingqiu and Chen, Guo and Zhang, Yuejie and Feng, Rui and Xie, Weidi},
  booktitle={ICLR},
  year={2025}
}

@inproceedings{zhang2026exo2ego,
  title={{Exo2Ego:} {Exocentric} Knowledge Guided {MLLM} for Egocentric Video Understanding},
  author={Zhang, Haoyu and Chu, Qiaohui and Liu, Meng and Shi, Haoxiang and Wang, Yaowei and Nie, Liqiang},
  booktitle={AAAI},
  year={2026}
}

@inproceedings{hu2025robust_ego_exo,
  title={Robust Ego-Exo Correspondence with Long-Term Memory},
  author={Hu, Yijun and Fan, Bing and Gu, Xin and Ren, Haiqing and Liu, Dongfang and Fan, Heng and Zhang, Libo},
  booktitle={NeurIPS},
  year={2025}
}

@inproceedings{huang2024egoexolearn,
  title={{EgoExoLearn:} {A} Dataset for Bridging Asynchronous Ego-and Exo-centric View of Procedural Activities in Real World},
  author={Huang, Yifei and Chen, Guo and Xu, Jilan and Zhang, Mingfang and Yang, Lijin and Pei, Baoqi and Zhang, Hongjie and Dong, Lu and Wang, Yali and Wang, Limin and Yu, Qiao},
  booktitle={CVPR},
  year={2024}
}

@inproceedings{mur2025omama,
  title={{O-MaMa:} {Learning} Object Mask Matching between Egocentric and Exocentric Views},
  author={Mur-Labadia, Lorenzo and Santos-Villafranca, Maria and Bermudez-Cameo, Jesus and Perez-Yus, Alejandro and Martinez-Cantin, Ruben and Guerrero, Jose J.},
  booktitle={ICCV},
  year={2025}
}

@article{hu2025cascaded_dynamic_memory,
  title={Cascaded dynamic memory refinement and semantic alignment for exo-to-ego cross-view video generation},
  author={Hu, Weipeng and Hoe, Jiun Tian and Li, Jianhui and Hu, Haifeng and Jiang, Xudong and Tan, Yap-Peng},
  journal={IEEE Transactions on Pattern Analysis and Machine Intelligence},
  year={2025},
  publisher={IEEE}
}

@inproceedings{lu2025ego_exo_viewpoint,
  title={Ego vs. Exo and Active vs. Passive: Investigating the Individual and Combined Effects of Viewpoint and Navigation on Spatial Immersion and Understanding in Immersive Storytelling},
  author={Lu, Tao and Zhu, Qian and Ma, Tiffany and Kam-Kwai, Wong and Xie, Anlan and Endert, Alex and Yang, Yalong},
  booktitle={CHI},
  year={2025}
}

@inproceedings{shi2026test_action_anticipation,
  title={Test-time Ego-Exo-centric Adaptation for Action Anticipation via Multi-Label Prototype Growing and Dual-Clue Consistency},
  author={Shi, Zhaofeng and Qiu, Heqian and Wang, Lanxiao and Wu, Qingbo and Meng, Fanman and Pan, Lili and Li, Hongliang},
  booktitle={CVPR},
  year={2026}
}

@article{wang2026towards_explainable,
  title={Towards top-down reasoning: An explainable multi-agent approach for visual question answering},
  author={Wang, Zeqing and Wan, Wentao and Lao, Qiqing and Chen, Runmeng and Lang, Minjie and Wang, Xiao and Gao, Feng and Wang, Keze and Lin, Liang},
  journal={IEEE Transactions on Multimedia},
  year={2026},
  publisher={IEEE}
}

@article{wang2026dual_space_intervention,
  title={Dual-Space Intervention for Mitigating Bias in Robust Visual Question Answering},
  author={Wang, Runmin and Song, Xingdong and Wan, Zukun and Xu, Han and Yu, Congzhen and Ma, Tianming and Ding, Yajun and Qian, Shengyou},
  journal={Expert Systems with Applications},
  year={2026},
  publisher={Elsevier}
}

@inproceedings{li2026egocross,
  title={{EgoCross:} {Benchmarking} Multimodal Large Language Models for Cross-Domain Egocentric Video Question Answering},
  author={Li, Yanjun and Fu, Yuqian and Qian, Tianwen and Xu, Qi'Ao and Dai, Silong and Paudel, Danda Pani and Van Gool, Luc and Wang, Xiaoling},
  booktitle={AAAI},
  year={2026}
}

@inproceedings{park2025bootstrap,
  title={Bootstrap Your Own Views: Masked Ego-Exo Modeling for Fine-grained View-invariant Video Representations},
  author={Park, Jungin and Lee, Jiyoung and Sohn, Kwanghoon},
  booktitle={CVPR},
  year={2025}
}

@article{li2026collaborated_hallucination,
  title={Collaborated with Hallucination: Enhancing Egocentric Grounded Question Answering via Error Demonstrations},
  author={Li, Shenshen and Xu, Xing and Shen, Fumin and Sun, Zhe and Cichocki, Andrzej and Shen, Heng Tao},
  journal={IEEE Transactions on Image Processing},
  year={2026},
  publisher={IEEE}
}

@inproceedings{yang2025egolife,
  title={{EgoLife:} {Towards} Egocentric Life Assistant},
  author={Jingkang Yang and
                  Shuai Liu and
                  Hongming Guo and
                  Yuhao Dong and
                  Xiamengwei Zhang and
                  Sicheng Zhang and
                  Pengyun Wang and
                  Zitang Zhou and
                  Binzhu Xie and
                  Ziyue Wang and
                  Bei Ouyang and
                  Zhengyu Lin and
                  Marco Cominelli and
                  Zhongang Cai and
                  Bo Li and
                  Yuanhan Zhang and
                  Peiyuan Zhang and
                  Fangzhou Hong and
                  Joerg Widmer and
                  Francesco Gringoli and
                  Lei Yang and
                  Ziwei Liu},
  booktitle={CVPR},
  year={2025}
}

@inproceedings{zhou2025egotextvqa,
  title={{EgoTextVQA:} {Towards} Egocentric Scene-Text Aware Video Question Answering},
  author={Zhou, Sheng and Xiao, Junbin and Li, Qingyun and Li, Yicong and Yang, Xun and Guo, Dan and Wang, Meng and Chua, Tat-Seng and Yao, Angela},
  booktitle={CVPR},
  year={2025}
}

@inproceedings{zhang2025egonight,
  title={{EgoNight:} {Towards} Egocentric Vision Understanding at Night with a Challenging Benchmark},
  author={Zhang, Deheng and Fu, Yuqian and Yang, Runyi and Miao, Yang and Qian, Tianwen and Zheng, Xu and Sun, Guolei and Chhatkuli, Ajad and Huang, Xuanjing and Jiang, Yu-Gang and Van Gool, Luc and Paudel, Danda Pani},
  booktitle={ICLR},
  year={2026}
}

@inproceedings{bai2023glance_focus,
  title={Glance and Focus: Memory Prompting for Multi-Event Video Question Answering},
  author={Bai, Ziyi and Wang, Ruiping and Chen, Xilin},
  booktitle={NeurIPS},
  year={2023}
}

@inproceedings{park2025egoworld,
  title={{EgoWorld:} {Translating} Exocentric View to Egocentric View using Rich Exocentric Observations},
  author={Park, Junho and Ye, Andrew Sangwoo and Kwon, Taein},
  booktitle={ICLR},
  year={2026}
}

@inproceedings{jiang2023single_stage,
  title={Single-stage visual query localization in egocentric videos},
  author={Jiang, Hanwen and Ramakrishnan, Santhosh Kumar and Grauman, Kristen},
  booktitle={NeurIPS},
  year={2023}
}

@inproceedings{jia2022egotaskqa,
  title={{EgoTaskQA:} {Understanding} Human Tasks in Egocentric Videos},
  author={Jia, Baoxiong and Lei, Ting and Zhu, Song-Chun and Huang, Siyuan},
  booktitle={NeurIPS},
  year={2022}
}

@inproceedings{pei2025egothinker,
  title={{EgoThinker:} {Unveiling} Egocentric Reasoning with Spatio-Temporal {CoT}},
  author={Pei, Baoqi and Huang, Yifei and Xu, Jilan and He, Yuping and Chen, Guo and Wu, Fei and Pang, Jiangmiao and Qiao, Yu},
  booktitle={NeurIPS},
  year={2025}
}

@inproceedings{ravi2025out_sight_context,
  title={Out of sight, not out of context? {Egocentric} spatial reasoning in vlms across disjoint frames},
  author={Ravi, Sahithya and Sarch, Gabriel Herbert and Vineet, Vibhav and Wilson, Andrew D. and Kumaravel, Balasaravanan Thoravi},
  booktitle={EMNLP},
  year={2025}
}

@inproceedings{xiaoegoblind,
  title={{EgoBlind:} {Towards} Egocentric Visual Assistance for the Blind},
  author={Xiao, Junbin and Huang, Nanxin and Qiu, Hao and Tao, Zhulin and Yang, Xun and Hong, Richang and Wang, Meng and Yao, Angela},
  booktitle={NeurIPS},
  year={2025}
}

@article{zhou2025scene_text_grounding,
  title={Scene-text grounding for text-based video question answering},
  author={Zhou, Sheng and Xiao, Junbin and Yang, Xun and Song, Peipei and Guo, Dan and Yao, Angela and Wang, Meng and Chua, Tat-Seng},
  journal={IEEE Transactions on Multimedia},
  year={2025},
  publisher={IEEE}
}

@inproceedings{xiao2026ego_grounding,
  title={Ego-Grounding for Personalized Question-Answering in Egocentric Videos},
  author={Xiao, Junbin and Zhang, Shenglang and Zhu, Pengxiang and Yao, Angela},
  booktitle={CVPR},
  year={2026}
}

@inproceedings{zamiri2026say_conversational,
  title={Say It My Way: Exploring Control in Conversational Visual Question Answering with Blind Users},
  author={Zamiri Zeraati, Farnaz and Cao, Yang and Qiao, Yuehan and Daum{\'e} III, Hal and Kacorri, Hernisa},
  booktitle={CHI},
  year={2026}
}

@inproceedings{sun2025visual_intention,
  title={Visual intention grounding for egocentric assistants},
  author={Sun, Pengzhan and Xiao, Junbin and Tse, Tze Ho Elden and Li, Yicong and Akula, Arjun and Yao, Angela},
  booktitle={ICCV},
  year={2025}
}

@inproceedings{wang2025object_centric_vqa,
  title={Object-centric video question answering with visual grounding and referring},
  author={Wang, Haochen and Chen, Qirui and Yan, Cilin and Cai, Jiayin and Jiang, Xiaolong and Hu, Yao and Xie, Weidi and Gavves, Stratis},
  booktitle={ICCV},
  year={2025}
}

@inproceedings{dang2025ecbench,
  title={{ECBench:} {Can} Multi-modal Foundation Models Understand the Egocentric World? {A} Holistic Embodied Cognition Benchmark},
  author={Dang, Ronghao and Yuan, Yuqian and Zhang, Wenqi and Xin, Yifei and Zhang, Boqiang and Li, Long and Wang, Liuyi and Zeng, Qinyang and Li, Xin and Bing, Lidong},
  booktitle={CVPR},
  year={2025}
}

@inproceedings{ragusa2026ego_extra,
  title={{Ego-EXTRA:} {video-language} Egocentric Dataset for EXpert-TRAinee assistance},
  author={Ragusa, Francesco and Mazzamuto, Michele and Forte, Rosario and D'Ambra, Irene and Fort, James and Engel, Jakob and Furnari, Antonino and Farinella, Giovanni Maria},
  booktitle={WACV},
  year={2026}
}

@inproceedings{peng2025eye,
  title={In the Eye of {MLLM}: {Benchmarking} Egocentric Video Intent Understanding with Gaze-Guided Prompting},
  author={Peng, Taiying and Hua, Jiacheng and Liu, Miao and Lu, Feng},
  booktitle={NeurIPS},
  year={2025}
}

@inproceedings{rodin2025easg,
  title={{EASG-Bench:} {Video} {Q\&A} Benchmark with Egocentric Action Scene Graphs},
  author={Rodin, Ivan and Wu, Tz-Ying and Min, Kyle and Sridhar, Sharath Nittur and Furnari, Antonino and Tripathi, Subarna and Farinella, Giovanni Maria},
  booktitle={ICCVW},
  year={2025}
}

@inproceedings{li2026learning_situated_awareness,
  title={Learning Situated Awareness in the Real World},
  author={Li, Chuhan and Han, Ruilin and Hsu, Joy and Liang, Yongyuan and Dhawan, Rajiv and Wu, Jiajun and Yang, Ming-Hsuan and Wang, Xin Eric},
  booktitle={ICML},
  year={2026}
}

@inproceedings{cherian2022_spatio_temporal_scene_graph,
  title={{(2.5+ 1) D} Spatio-Temporal Scene Graphs for Video Question Answering},
  author={Cherian, Anoop and Hori, Chiori and Marks, Tim K and Le Roux, Jonathan},
  booktitle={AAAI},
  year={2022}
}

@inproceedings{sima2024drivelm,
  title={{DriveLM:} {Driving} with Graph Visual Question Answering},
  author={Sima, Chonghao and Renz, Katrin and Chitta, Kashyap and Chen, Li and Zhang, Hanxue and Xie, Chengen and Bei{\ss}wenger, Jens and Luo, Ping and Geiger, Andreas and Li, Hongyang},
  booktitle={ECCV},
  year={2024}
}

@inproceedings{li2026mllms_referential,
  title={Do {MLLMs} Understand Pointing? {Benchmarking} and Enhancing Referential Reasoning in Egocentric Vision},
  author={Li, Chentao and Gao, Zirui and Gao, Mingze and Ren, Yinglian and Feng, Jianjiang and Zhou, Jie},
  booktitle={ACL},
  year={2026}
}

@inproceedings{luo2025openmmego,
  title={{OpenMMEgo:} {Enhancing} egocentric understanding for {LMMs} with open weights and data},
  author={Luo, Hao and Yue, Zihao and Zhang, Wanpeng and Feng, Yicheng and Zheng, Sipeng and Ye, Deheng and Lu, Zongqing},
  booktitle={NeurIPS},
  year={2025}
}

@inproceedings{jung2025right,
  title={Is `right' right? {Enhancing} object orientation understanding in multimodal large language models through egocentric instruction tuning},
  author={Jung, Ji Hyeok and Kim, Eun Tae and Kim, Seoyeon and Lee, Joo Ho and Kim, Bumsoo and Chang, Buru},
  booktitle={CVPR},
  year={2025}
}

@inproceedings{yuan2025walkvlm,
  title={{WalkVLM:} {Aid} Visually Impaired People Walking by Vision Language Model},
  author={Yuan, Zhiqiang and Zhang, Ting and Zhu, Yeshuang and Zhang, Jiapei and Deng, Ying and Jia, Zexi and Luo, Peixiang and Duan, Xiaoyue and Zhou, Jie and Zhang, Jinchao},
  booktitle={ICCV},
  year={2025}
}

@inproceedings{sultan2026walkgpt,
  title={{WalkGPT:} {Grounded} Vision-Language Conversation with Depth-Aware Segmentation for Pedestrian Navigation},
  author={Sultan, Rafi Ibn and Zhu, Hui and Zhou, Xiangyu and Li, Chengyin and Khanduri, Prashant and Brocanelli, Marco and Zhu, Dongxiao},
  booktitle={CVPR},
  year={2026}
}

@inproceedings{tang2025adaptive,
  title={Adaptive keyframe sampling for long video understanding},
  author={Tang, Xi and Qiu, Jihao and Xie, Lingxi and Tian, Yunjie and Jiao, Jianbin and Ye, Qixiang},
  booktitle={CVPR},
  year={2025}
}

@inproceedings{liu2025bolt,
  title={{BOLT:} {Boost} Large Vision-Language Model Without Training for Long-form Video Understanding},
  author={Liu, Shuming and Zhao, Chen and Xu, Tianqi and Ghanem, Bernard},
  booktitle={CVPR},
  year={2025}
}

@inproceedings{jeong2025videorag,
  title={{VideoRAG:} {Retrieval-augmented} generation over video corpus},
  author={Jeong, Soyeong and Kim, Kangsan and Baek, Jinheon and Hwang, Sung Ju},
  booktitle={ACL (Findings)},
  year={2025}
}

@inproceedings{luo2024video,
  title={{Video-RAG:} {Visually-aligned} retrieval-augmented long video comprehension},
  author={Yongdong Luo and
                  Xiawu Zheng and
                  Xiao Yang and
                  Guilin Li and
                  Haojia Lin and
                  Jinfa Huang and
                  Jiayi Ji and
                  Fei Chao and
                  Jiebo Luo and
                  Rongrong Ji},
  booktitle={NeurIPS},
  year={2025}
}

@article{kim2026ma,
  title={{MA-EgoQA:} {Question} Answering over Egocentric Videos from Multiple Embodied Agents},
  author={Kim, Kangsan and Yang, Yanlai and Kim, Suji and Yeo, Woongyeong and Lee, Youngwan and Ren, Mengye and Hwang, Sung Ju},
  journal={arXiv preprint arXiv:2603.09827},
  year={2026}
}

@inproceedings{yeo2025worldmm,
  title={{WorldMM:} {Dynamic} Multimodal Memory Agent for Long Video Reasoning},
  author={Yeo, Woongyeong and Kim, Kangsan and Yoon, Jaehong and Hwang, Sung Ju},
  booktitle={CVPR},
  year={2026}
}

@article{robertson2009bm25,
author = {Robertson, Stephen and Zaragoza, Hugo},
title = {The Probabilistic Relevance Framework: {BM25} and Beyond},
year = {2009},
journal = {Information Retrieval}
}

@inproceedings{karpukhin2020dense,
  title={Dense passage retrieval for open-domain question answering},
  author={Karpukhin, Vladimir and Oguz, Barlas and Min, Sewon and Lewis, Patrick and Wu, Ledell and Edunov, Sergey and Chen, Danqi and Yih, Wen-tau},
  booktitle={EMNLP},
  year={2020}
}

@article{chen2026egoplan,
  title={{EgoPlan-Bench:} {Benchmarking} Multimodal Large Language Models for Human-Level Planning},
  author={Chen, Yi and Ge, Yuying and Ge, Yixiao and Ding, Mingyu and Li, Bohao and Wang, Rui and Xu, Ruifeng and Shan, Ying and Liu, Xihui},
  journal={International Journal of Computer Vision},
  year={2026},
  publisher={Springer}
}

@inproceedings{yang2025thinking,
  title={Thinking in space: How multimodal large language models see, remember, and recall spaces},
  author={Yang, Jihan and Yang, Shusheng and Gupta, Anjali W and Han, Rilyn and Fei-Fei, Li and Xie, Saining},
  booktitle={CVPR},
  year={2025}
}

@inproceedings{mangalam2023egoschema,
  title={{EgoSchema:} {A} Diagnostic Benchmark for Very Long-form Video Language Understanding},
  author={Mangalam, Karttikeya and Akshulakov, Raiymbek and Malik, Jitendra},
  booktitle={NeurIPS},
  year={2023}
}

@inproceedings{ye2025mmego,
  title={{MM-Ego:} {Towards} Building Egocentric Multimodal {LLMs} for Video {QA}},
  author={Ye, Hanrong and Zhang, Haotian and Daxberger, Erik and Chen, Lin and Lin, Zongyu and Li, Yanghao and Zhang, Bowen and You, Haoxuan and Xu, Dan and Gan, Zhe and Lu, Jiasen and Yang, Yinfei},
  booktitle={ICLR},
  year={2025}
}

@inproceedings{he2025egoexobench,
  title={{EgoExoBench:} {A} Benchmark for First-and Third-person View Video Understanding in {MLLMs}},
  author={He, Yuping and Huang, Yifei and Chen, Guo and Pei, Baoqi and Xu, Jilan and Lu, Tong and Pang, Jiangmiao},
  booktitle={NeurIPS},
  year={2025}
}

@article{endsley1995toward,
  title={Toward a Theory of Situation Awareness in Dynamic Systems},
  author={Endsley, Mica R.},
  journal={Human Factors: The Journal of the Human Factors and Ergonomics Society},
  year={1995}
}

@article{nigro1983pov,
title = {Point of view in personal memories},
journal = {Cognitive Psychology},
year = {1983},
author = {Georgia Nigro and Ulric Neisser}
}

@article{burgess2006spatial,
title = {Spatial memory: how egocentric and allocentric combine},
journal = {Trends in Cognitive Sciences},
year = {2006},
author = {Neil Burgess},
}

@inproceedings{majumdar2023openeqa,
    author = {Arjun Majumdar and
                  Anurag Ajay and
                  Xiaohan Zhang and
                  Pranav Putta and
                  Sriram Yenamandra and
                  Mikael Henaff and
                  Sneha Silwal and
                  Paul McVay and
                  Oleksandr Maksymets and
                  Sergio Arnaud and
                  Karmesh Yadav and
                  Qiyang Li and
                  Ben Newman and
                  Mohit Sharma and
                  Vincent{-}Pierre Berges and
                  Shiqi Zhang and
                  Pulkit Agrawal and
                  Yonatan Bisk and
                  Dhruv Batra and
                  Mrinal Kalakrishnan and
                  Franziska Meier and
                  Chris Paxton and
                  Alexander Sax and
                  Aravind Rajeswaran},
  title={{OpenEQA:} {Embodied} Question Answering in the Era of Foundation Models},
  booktitle={CVPR},
  year={2024},
}

@inproceedings{grauman2024ego,
  title={{Ego-Exo4D:} {Understanding} skilled human activity from first-and third-person perspectives},
  author={Grauman, Kristen and Westbury, Andrew and Torresani, Lorenzo and Kitani, Kris and Malik, Jitendra and Afouras, Triantafyllos and Ashutosh, Kumar and Baiyya, Vijay and Bansal, Siddhant and Boote, Bikram and others},
  booktitle={CVPR},
  year={2024}
}

@inproceedings{jia2020lemma,
  title={{LEMMA:} {A} Multi-view Dataset for Learning Multi-agent Multi-task Activities},
  author={Jia, Baoxiong and Chen, Yixin and Huang, Siyuan and Zhu, Yixin and Zhu, Song-Chun},
  booktitle={ECCV},
  year={2020}
}

@inproceedings{chen2026mural,
  title = {{MuRAL:} {A} Multi-Resident Ambient Sensor Dataset Annotated with Natural Language for Activities of Daily Living},
  author = {Chen, Xi and Cumin, Julien and Ramparany, Fano and Vaufreydaz, Dominique},
  booktitle = {ICIE},
  year = {2026}
}

@article{schneider2025omnifall,
  title={{OmniFall:} {A} Unified Staged-to-Wild Benchmark for Human Fall Detection},
  author={Schneider, David and Marinov, Zdravko and Baur, Rafael and Zhong, Zeyun and D{\"u}ger, Rodi and Stiefelhagen, Rainer},
  journal={arXiv preprint arXiv:2505.19889},
  year={2025}
}

@article{gabriel2025continuous,
  title={Continuous patient monitoring with {AI}: {Real-time} analysis of video in hospital care settings},
  author={Gabriel, Paolo and Rehani, Peter and Troy, Tyler and Wyatt, Tiffany and Choma, Michael and Singh, Narinder},
  journal={Frontiers in Imaging},
  year={2025},
  publisher={Frontiers Media SA}
}

@misc{ring2024,
  title = {Ring Home Security Systems},
  howpublished = {\url{https://ring.com}},
  year = {2024}
}

@inproceedings{zhou2025learning,
  title={Learning {3D} Persistent Embodied World Models},
  author={Zhou, Siyuan and Du, Yilun and Yang, Yuncong and Han, Lei and Chen, Peihao and Yeung, Dit-Yan and Gan, Chuang},
  booktitle={NeurIPS},
  year={2025}
}

@article{liu2024objectfinder,
  title={{ObjectFinder:} {An} Open-Vocabulary Assistive System for Interactive Object Search by Blind People},
  author={Liu, Ruiping and Zhang, Jiaming and Sch{\"o}n, Angela and M{\"u}ller, Karin and Zheng, Junwei and Yang, Kailun and Guo, Anhong and Gerling, Kathrin and Stiefelhagen, Rainer},
  journal={arXiv preprint arXiv:2412.03118},
  year={2024}
}

@article{comanici2025gemini,
  title={Gemini 2.5: Pushing the frontier with advanced reasoning, multimodality, long context, and next generation agentic capabilities},
  author={Comanici, Gheorghe and Bieber, Eric and Schaekermann, Mike and Pasupat, Ice and Sachdeva, Noveen and Dhillon, Inderjit and Blistein, Marcel and Ram, Ori and Zhang, Dan and Rosen, Evan and others},
  journal={arXiv preprint arXiv:2507.06261},
  year={2025}
}

@article{wang2025internvl3,
  title={{InternVL3.5:} {Advancing} Open-Source Multimodal Models in Versatility, Reasoning, and Efficiency},
  author={Wang, Weiyun and Gao, Zhangwei and Gu, Lixin and Pu, Hengjun and Cui, Long and Wei, Xingguang and Liu, Zhaoyang and Jing, Linglin and Ye, Shenglong and Shao, Jie and others},
  journal={arXiv preprint arXiv:2508.18265},
  year={2025}
}

@article{li2024llava,
  title={{LLaVA-OneVision:} {Easy} visual task transfer},
  author={Li, Bo and Zhang, Yuanhan and Guo, Dong and Zhang, Renrui and Li, Feng and Zhang, Hao and Zhang, Kaichen and Zhang, Peiyuan and Li, Yanwei and Liu, Ziwei and others},
  journal={arXiv preprint arXiv:2408.03326},
  year={2024}
}

@article{bai2025qwen3,
  title={{Qwen3-VL} technical report},
  author={Bai, Shuai and Cai, Yuxuan and Chen, Ruizhe and Chen, Keqin and Chen, Xionghui and Cheng, Zesen and Deng, Lianghao and Ding, Wei and Gao, Chang and Ge, Chunjiang and others},
  journal={arXiv preprint arXiv:2511.21631},
  year={2025}
}

@article{li2024llava_next_interleave,
  title={{LLaVA-NeXT-Interleave:} {Tackling} Multi-image, Video, and {3D} in Large Multimodal Models},
  author={Li, Feng and Zhang, Renrui and Zhang, Hao and Zhang, Yuanhan and Li, Bo and Li, Wei and Ma, Zejun and Li, Chunyuan},
  journal={arXiv preprint arXiv:2407.07895},
  year={2024}
}

@inproceedings{xue2025adavideorag,
  title={{AdaVideoRAG:} {Omni-contextual} Adaptive Retrieval-Augmented Efficient Long Video Understanding},
  author={Xue, Zhucun and Zhang, Jiangning and Xie, Xurong and Cai, Yuxuan and Liu, Yong and Li, Xiangtai and Tao, Dacheng},
  booktitle={NeurIPS},
  year={2025}
}

@inproceedings{kulesza2011kdpps,
author = {Kulesza, Alex and Taskar, Ben},
title = {{k-DPPs:} {Fixed-size} Determinantal Point Processes},
year = {2011},
booktitle = {ICML}
}

@inproceedings{yang2024egoposeformer,
  title={{EgoPoseFormer:} {A} simple baseline for stereo egocentric {3D} human pose estimation},
  author={Chenhongyi Yang and
                  Anastasia Tkach and
                  Shreyas Hampali and
                  Linguang Zhang and
                  Elliot J. Crowley and
                  Cem Keskin},
  booktitle={ECCV},
  year={2024}
}

@inproceedings{grauman2022ego4d,
  title={{Ego4D:} {Around} the World in 3,000 Hours of Egocentric Video},
  author={Grauman, Kristen and Westbury, Andrew and Byrne, Eugene and Chavis, Zachary and Furnari, Antonino and Girdhar, Rohit and Hamburger, Jackson and Jiang, Hao and Liu, Miao and Liu, Xingyu and others},
  booktitle={CVPR},
  year={2022}
}

@inproceedings{fan2025embodied,
  title={Embodied {VideoAgent}: {Persistent} Memory from Egocentric Videos and Embodied Sensors Enables Dynamic Scene Understanding},
  author={Fan, Yue and Ma, Xiaojian and Su, Rongpeng and Guo, Jun and Wu, Rujie and Chen, Xi and Li, Qing},
  booktitle={ICCV},
  year={2025}
}

@article{yadav2024findingdory,
  title={{FindingDory:} {A} Benchmark to Evaluate Memory in Embodied Agents},
  author={Yadav, Karmesh and Ali, Yusuf and Gupta, Gunshi and Gal, Yarin and Kira, Zsolt},
  journal={arXiv preprint arXiv:2506.15635},
  year={2025}
}

@article{liu2025aligning,
  title={Aligning cyber space with physical world: A comprehensive survey on embodied {AI}},
  author={Liu, Yang and Chen, Weixing and Bai, Yongjie and Liang, Xiaodan and Li, Guanbin and Gao, Wen and Lin, Liang},
  journal={IEEE/ASME Transactions on Mechatronics},
  year={2025},
  publisher={IEEE}
}

@article{jung2025egoexo,
  title={{EgoExo-Con:} {Exploring} View-Invariant Video Temporal Understanding},
  author={Jung, Minjoon and Xiao, Junbin and Kim, Junghyun and Zhang, Byoung-Tak and Yao, Angela},
  journal={arXiv preprint arXiv:2510.26113},
  year={2025}
}

@article{reilly2025my,
  title={From My View to Yours: Ego-to-Exo Transfer in VLMs for Understanding Activities of Daily Living},
  author={Reilly, Dominick and Govind, Manish Kumar and Xue, Le and Das, Srijan},
  journal={arXiv preprint arXiv:2501.05711},
  year={2025}
}

@inproceedings{chen2025savvy,
    title={{SAVVY:} {Spatial} Awareness via Audio-Visual {LLMs} through Seeing and Hearing},
    author={Mingfei Chen and Zijun Cui and Xiulong Liu and Jinlin Xiang and Caleb Zheng and Jingyuan Li and Eli Shlizerman},
    year={2025},
    booktitle={NeurIPS}
}

@misc{openai2026gpt54,
  title        = {Introducing {GPT-5.4}},
  author       = {{OpenAI}},
  year         = {2026},
  month        = mar,
  day          = {5},
  howpublished = {\url{https://openai.com/index/introducing-gpt-5-4}},
  note         = {Accessed: 2026-05-05}
}

@inproceedings{radford2021learning,
  title={Learning transferable visual models from natural language supervision},
  author={Alec Radford and
                  Jong Wook Kim and
                  Chris Hallacy and
                  Aditya Ramesh and
                  Gabriel Goh and
                  Sandhini Agarwal and
                  Girish Sastry and
                  Amanda Askell and
                  Pamela Mishkin and
                  Jack Clark and
                  Gretchen Krueger and
                  Ilya Sutskever},
  booktitle={ICML},
  year={2021}
}

@misc{openai2024gpt4o,
  title        = {Hello {GPT-4o}},
  author       = {{OpenAI}},
  year         = {2024},
  month        = may,
  day          = {13},
  howpublished = {\url{https://openai.com/index/hello-gpt-4o}},
  note         = {Accessed: 2026-05-05}
}

@inproceedings{derezinski2019exact,
  title={Exact sampling of determinantal point processes with sublinear time preprocessing},
  author={Derezinski, Michal and Calandriello, Daniele and Valko, Michal},
  booktitle={NeurIPS},
  year={2019}
}

@inproceedings{fu2025video,
  title={Video-MME: The first-ever comprehensive evaluation benchmark of multi-modal llms in video analysis},
  author={Fu, Chaoyou and Dai, Yuhan and Luo, Yongdong and Li, Lei and Ren, Shuhuai and Zhang, Renrui and Wang, Zihan and Zhou, Chenyu and Shen, Yunhang and Zhang, Mengdan and others},
  booktitle={CVPR},
  year={2025}
}

@article{liu2025screens,
  title={From screens to scenes: A survey of embodied AI in healthcare},
  author={Liu, Yihao and Cao, Xu and Chen, Tingting and Jiang, Yankai and You, Junjie and Wu, Minghua and Wang, Xiaosong and Feng, Mengling and Jin, Yaochu and Chen, Jintai},
  journal={Information Fusion},
  volume={119},
  pages={103033},
  year={2025},
  publisher={Elsevier}
}

@article{bai2025qwen25vl,
  title={{Qwen2.5-VL} Technical Report},
  author={Bai, Shuai and Chen, Keqin and Liu, Xuejing and Wang, Jialin and Ge, Wenbin and Song, Sibo and Dang, Kai and Wang, Peng and Wang, Shijie and Tang, Jun and others},
  journal={arXiv preprint arXiv:2502.13923},
  year={2025}
}

\newpage
\appendix
\section{Social Impact and Limitations}
\label{sec:limitation}
Memory is essential for reasoning with MLLMs as a form of auxiliary cognition. However, existing memory-based approaches rely solely on the egocentric stream of mobile agents and their interactions with the environment. This limits their ability to capture full-body movements of other agents and track their interactions with the environment, both of which are critical to the holistic understanding of the scene.
With the advancement of surveillance systems, wearable devices for human agents, and visual sensors for embodied agents, memory can increasingly be constructed from cross-view sources to enable more comprehensive retrieval, \textit{e.g.}, in homes~\cite{jia2020lemma}, hospitals~\cite{liu2025screens}, and parking facilities~\cite{he2026bridging}.

As the first benchmark for ego-exo memory, EgoExoMem has several limitations. First, the minute-level video duration limits the scope of our benchmark, where simple retrieval methods may already suffice. Since real-world observations can span weeks or months, datasets supporting long-term ego-exo memory are necessary, and structured retrieval methods could prove more beneficial in such settings. Second, to standardize the task, we currently restrict input to one egocentric and one exocentric stream. However, the underlying datasets contain multiple egocentric and exocentric views, which means that the same MCQs could be repurposed for multi-view settings. Future work could explore this direction and investigate how multiple egocentric or exocentric streams individually contribute to benchmark performance.
\section{Human Editing and Filtering}
\begin{figure}[h]
    \centering
    \includegraphics[width=\linewidth]{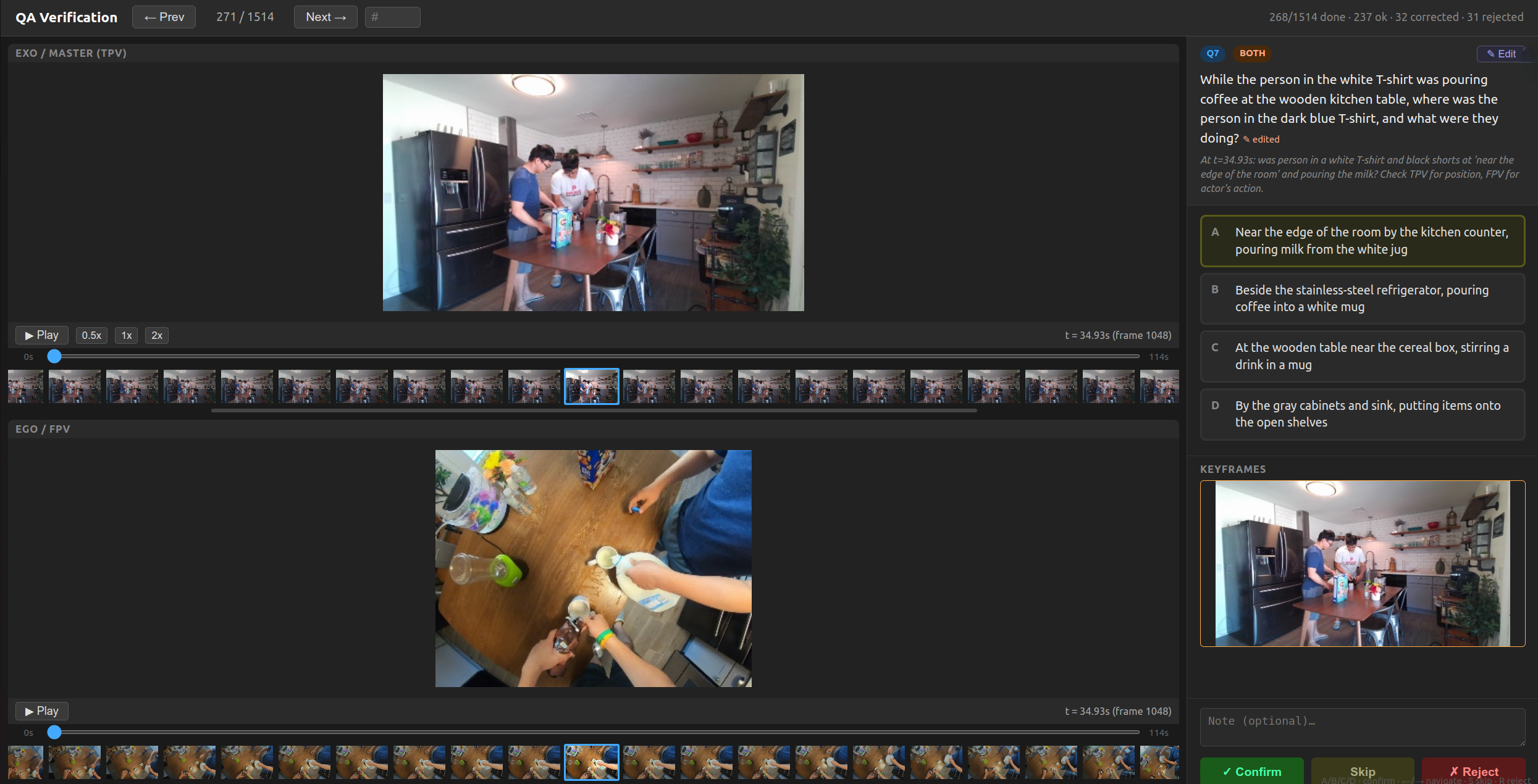}
    \caption{Verification tool for human annotator editing and filtering.}
    \label{fig:verification}
\end{figure}

Fig.~\ref{fig:verification} shows the verification user interface.
\section{Evaluation Prompts}
\label{sec:prompts}
The prompt used to generate captions for RAG-based methods with Gemini 2.5 Flash is shown in Fig.~\ref{fig:cap_gen}. The evaluation template is provided in Fig.~\ref{fig:eval_temp}.
\begin{figure}[t]
\centering
\begin{tcolorbox}
\fontsize{8}{10.5}\selectfont

\textcolor{blue}{system\_prompt} =
\begin{quote}
\ttfamily
``You are a video captioning assistant. Given a few sampled frames from a short video clip, write ONE concise sentence describing the main activity or scene. Focus on actions and objects visible.''
\end{quote}
\end{tcolorbox}
\caption{Caption generation used for retrieval in RAG-based methods.}
\label{fig:cap_gen}
\end{figure}
\begin{figure}[t]
\centering
\begin{tcolorbox}
\fontsize{8}{10.5}\selectfont

\textcolor{blue}{system\_prompt} =
\begin{quote}
\ttfamily
``You are answering multiple-choice questions about a video. Selected keyframes from egocentric and exocentric views are provided, ordered by timestamp. Each frame is labeled with its view [Ego/Exo] and time. Answer with a single letter (A, B, C, or D) only. No explanation.''
\end{quote}

\textcolor{blue}{user\_prompt} =
\begin{quote}
\ttfamily
``Question: \{question\}\\
Options:\\[2pt]
A. \{option\_0\}\\
B. \{option\_1\}\\
C. \{option\_2\}\\
D. \{option\_3\}\\[2pt]
Answer with the option letter only (A, B, C, or D).''
\end{quote}

\end{tcolorbox}
\caption{Evaluation template.}
\label{fig:eval_temp}
\end{figure}
\section{Further Ablation Studies}
Following standard practice~\cite{fu2025video}, we uniformly sample $32$ frames as input. For RAG-based methods, we retrieve the top-$k$ clips ($k=8$) and sample $4$ frames per clip. An ablation study on the effect of $k$ is provided in Tab.~\ref{tab:topk}, which demonstrates that performance is largely insensitive to the choice of $k$.
\begin{table}[]
    \centering
    \caption{Ablation study on the effect of the top-$k$ value in VideoRAG~\cite{jeong2025videorag}.}
    \begin{tabular}{c|c|c|c|c|c|c|c|c|c|c}
    \toprule
       \textbf{Top-$k$} & \textbf{Views} & \textbf{HL} & \textbf{IP} & \textbf{RL} & \textbf{ED} & \textbf{OS} & \textbf{AR} & \textbf{TPA} & \textbf{TO} & \textbf{Avg} \\
    \midrule
\multirow{3}{*}{$k=5$}
& Ego     & 46.9 & 53.9 & 59.6 & 37.1 & 50.6 & 45.0 & 47.3 & 38.0 & 47.3 \\
& Exo     & 49.0 & 66.6 & 55.1 & 31.2 & 48.8 & 40.8 & 39.8 & 35.0 & 45.8 \\
\rowcolor{gray!20}
\cellcolor{white} & Ego+Exo & 53.5 & 65.8 & 60.9 & 33.2 & 54.3 & 46.7 & 41.9 & 39.9 & 49.5 \\
\midrule
\multirow{3}{*}{$k=8$}
& Ego     & 47.2 & 53.6 & 62.8 & 36.6 & 53.0 & 40.8 & 48.0 & 40.3 & 47.8 \\
& Exo     & 47.2 & 69.0 & 60.3 & 29.6 & 46.3 & 37.9 & 41.4 & 36.0 & 46.0 \\
\rowcolor{gray!20}
\cellcolor{white} & Ego+Exo & 49.0 & 66.6 & 62.2 & 33.5 & \textbf{57.9} & 43.8 & 41.0 & 39.6 & 49.2 \\
    \bottomrule
    \end{tabular}
    \label{tab:topk}
\end{table}


\end{document}